\documentclass[]{academic_template}

\usepackage[toc,page,header]{appendix}

\usepackage{amssymb}
\usepackage{amsmath}

\usepackage{makecell}
\usepackage{algorithm}      
\usepackage{algpseudocode}  

\usepackage{graphicx}

\usepackage{listings}
\usepackage{marvosym}

\usepackage{url}
\usepackage{booktabs}
\usepackage{wrapfig}

\newcommand{\best}[1]{\textbf{\scalebox{0.82}{#1}}}

\usepackage{colortbl}
\usepackage{xcolor}

\usepackage{caption}
\usepackage{subcaption}

\setlogoheight{14mm}   
\setlogospacing{5mm}
\setsjtublue 

\settitlerulethickness{3pt}  

\abstractboxon 
\setabstractframecolor{gray} 
\setabstractbgcolor{gray!10} 
\setlogotolineshift{5mm} 

\settoprulethickness{2.5pt}  
\setbottomrulethickness{1.5pt}

\title{Beyond Visual Enhancement: Adaptive Multi-Context Steering to Mitigate LVLM Hallucinations}

\author[1]{Shuran Ma}
\author[2]{JiaLe Li}
\author[3]{Yuxin Dong}
\author[3]{Shan Zheng}
\author[1]{Qingyun Jiang}
\author[4]{Xiang Chen}
\author[5]{Qi Zhu}
\author[5]{Deyi Ji}
\author[1]{Yifan Yang}
\author[]{Jianfeng Pan}
\author[6,\dagger]{Yu Tian}
\author[1,\mbox{\Letter}]{Xue Yang}

\affiliation[1]{Shanghai Jiao Tong University}
\affiliation[2]{Peking University}
\affiliation[3]{Beijing University of Chemical Technology}
\affiliation[4]{Nanjing University of Aeronautics and Astronautics}
\affiliation[5]{University of Science and Technology of China}
\affiliation[6]{Tsinghua University}
\contribution[\dagger]{Project Lead}
\contribution[\mbox{\Letter}]{Corresponding Author}

\abstract{
Hallucination remains a significant challenge in Large Vision-Language Models (LVLMs). Existing training-free methods generally mitigate hallucinations through contrastive decoding or visual enhancement, often increasing the relative influence of visual evidence during generation. This raises a fundamental question: \textbf{Can LVLMs dynamically regulate the contributions of different context sources to suppress hallucinations?} In this work, we investigate and quantify how LVLMs coordinate multiple context sources during decoding and examine how this intrinsic behavior can guide hallucination mitigation. We find that LVLMs exhibit an intrinsic vision-attending tendency that can guide adaptive visual steering, while textual contexts can also contribute to hallucination mitigation. Motivated by these findings, we propose \textbf{AIMS} (\underline{A}daptive \underline{I}nformation \underline{M}ulti-source \underline{S}teering), a lightweight training-free framework that adaptively coordinates visual, prefilled textual, and generated contexts during decoding. Specifically, AIMS constructs compact prototypes for the three context domains and estimates their affinities with the current query to determine head-wise steering weights. The resulting multi-source steering direction is applied to the query representation, enabling adaptive context integration without additional model training or auxiliary forward passes. Extensive experiments across multiple LVLMs and decoding strategies demonstrate that AIMS effectively mitigates object hallucination while maintaining competitive general-purpose multimodal capabilities. 
}

\date{\today}

\checkdata[Code]{\url{https://github.com/VisionXLab/AIMS}}
\checkdata[Hugging Face]{\url{https://huggingface.co/datasets/VisionXLab/AIMS_Benchmarks}}
\begin{document}
\maketitle


\section{Introduction}
Recent advances in Large Vision-Language Models (LVLMs)~\citep{chen2023shikra,llava1.5,peng2025lmm,minigpt4,NEURIPS2024_wings,NEURIPS2024_harmony} have enabled remarkable progress in bridging visual perception and language reasoning. By integrating powerful vision encoders with large language models (LLMs), LVLMs demonstrate strong capabilities in tasks such as image understanding~\citep{li2023blip}, visual question answering~\citep{lee2024visual,wang2024weakly}, and multimodal reasoning~\citep{deng2026openvlthinker,wang2026vl-rethinker,yang2025r1-onevision}.

Despite their remarkable capabilities, LVLMs still suffer from hallucination~\citep{liu2024survey,deng2024seeing}, generating content inconsistent with visual inputs, such as non-existent objects, incorrect attributes, or erroneous relationships. Unlike hallucination in text-only LLMs~\citep{shi2024trusting,huang2025survey}, LVLM hallucination is inherently tied to the interaction between visual evidence and language generation, posing a critical challenge to model reliability.

\begin{figure*}[t]
    \centering
    \includegraphics[width=\textwidth]{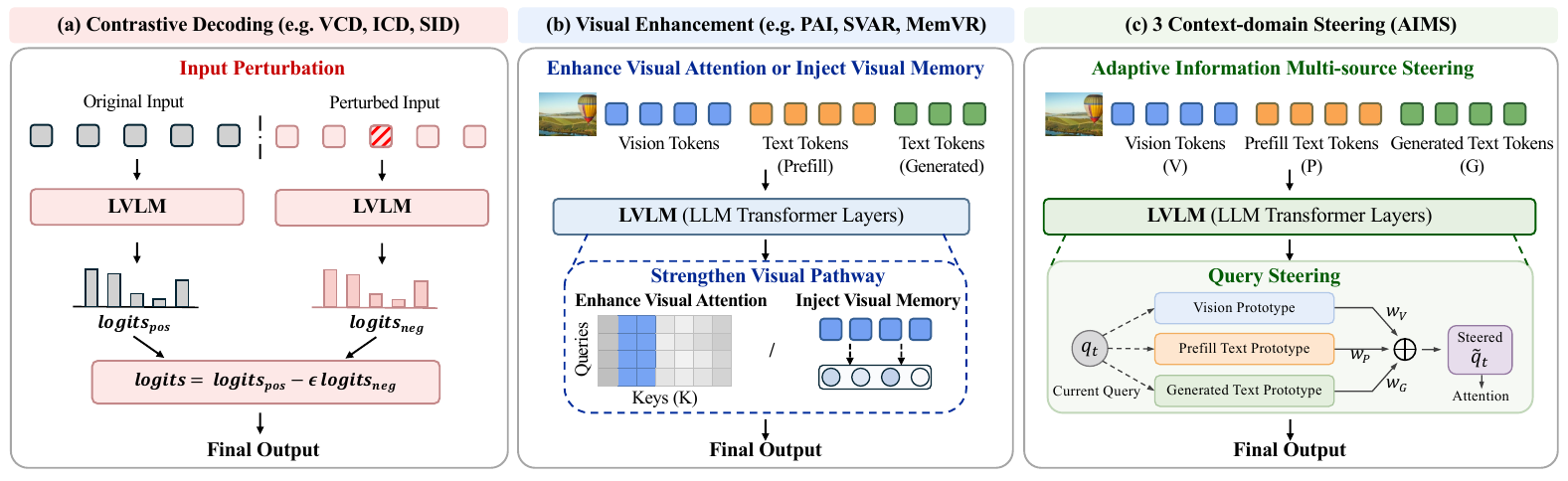}
    \caption{Comparison of training-free hallucination mitigation methods for LVLMs. From left to right: contrastive decoding, visual enhancement, and our method that dynamically steers the query toward three context branches during generation.}
    \label{fig:intro previous works}
\end{figure*}

As illustrated in Fig.\ref{fig:intro previous works}, existing training-free methods mainly mitigate hallucinations through contrastive decoding~\citep{VCD,ICD_LVLM,SID} or visual enhancement~\citep{PAI,SVAR_devils,memvr}. However, they largely overlook the dynamic utilization of contextual information during generation and focus predominantly on visual evidence, leaving textual prompts and generation history underexplored. This raises a fundamental question: \textbf{Can LVLMs dynamically regulate different context sources to suppress hallucinations?}

To better explore this issue, we design a series of preliminary experiments to quantify how LVLMs coordinate multiple context sources during autoregressive generation and examine how this intrinsic behavior can guide hallucination mitigation. Our analysis yields three observations. First, LVLMs exhibit a stage-dependent vision-attending tendency, with visual attention peaking during object grounding and decreasing as generation shifts toward semantic summarization. Second, directly exploiting this intrinsic tendency through query-adaptive steering mitigates hallucination more effectively than uniformly amplifying visual information. Third, prefilled text and generation history also provide corrective signals beyond visual evidence. Together, these findings highlight the need to adaptively coordinate multiple context sources rather than uniformly enhancing visual information.

Building on these observations, we propose \textbf{A}daptive \textbf{I}nformation \textbf{M}ulti-source \textbf{S}teering (AIMS), a lightweight training-free framework that adaptively coordinates visual context ($V$), prefilled text context ($P$), and previously generated context ($G$). Specifically, AIMS constructs semantic prototypes for the three contextual sources and estimates their query-dependent affinities at each decoding step. The affinity-weighted prototypes are aggregated to steer the current query representation, enabling dynamic multi-source coordination according to the generation state. Extensive experiments on four benchmarks demonstrate the effectiveness of AIMS, achieving up to a 27.1\% relative reduction in CHAIR $C_S$ across different LVLMs and decoding strategies, while preserving general multimodal capabilities. Our main contributions are summarized as follows:

\begin{itemize}
\item We investigate the contextual utilization behavior of LVLMs during autoregressive generation and reveal that visual reliance dynamically changes across generation stages. Our analysis further shows that hallucination mitigation requires considering multiple contextual sources beyond visual information alone.

\item We propose Adaptive Information Multi-source Steering (AIMS), a training-free framework that dynamically coordinates vision, prefilled text, and generated contexts through query-context affinity. Unlike existing methods that rely on fixed amplification of a single information source, AIMS adaptively integrates evidence during generation.

\item We conduct extensive experiments on multiple representative LVLMs, including LLaVA-1.5, Qwen2.5-VL, and Qwen3.5, across four widely used benchmarks. The results demonstrate that AIMS consistently reduces hallucination while effectively preserving overall generation quality compared with existing training-free approaches.
\end{itemize}

\section{Related Work}
\subsection{LARGE VISION-LANGUAGE MODELS}
Modern Large Vision-Language Models (LVLMs) have evolved rapidly through the scaling of architectural connections and language backbones. Early paradigms utilized LLaMA~\citep{llama, llama2} as the linguistic foundation, employing simple MLPs or Q-Formers to project visual tokens, as exemplified by LLaVA~\citep{llava, llava1.5} and MiniGPT series~\citep{minigpt4, minigptv2}. To handle denser multimodal semantics, subsequent architectures transitioned to highly optimized backbones; the Qwen-VL series~\citep{qwenvl, qwen2.5vl} introduced VIT-perceiver to internalize fine-grained visual features, while the InternLM-XComposer series~\citep{internlm_xcomposer, internlm_xcomposer2.5} and InternVL~\citep{internvl, internvl2.5} scaled up the LLM bases and integrated dynamic high-resolution vision-language interaction. Despite their superior multimodal capabilities, contemporary LVLMs still suffer from severe object hallucination problems~\citep{chair}. Effectively mitigating these hallucinations during inference remains a critical challenge, motivating our training-free dynamic steering framework.

\subsection{MITIGATING HALLUCINATIONS IN LVLMS}
In LVLMs, object hallucination remains the most prevalent form of hallucination, typically manifesting as errors in object category, attribute, and inter-object relation descriptions. To mitigate this issue, existing methods can be broadly divided into several lines. Early studies focused on improving fine-grained vision-language alignment and reducing co-occurrence bias in captioning models~\citep{chair}. More recent approaches leverage hallucination-oriented supervision, such as dedicated fine-tuning datasets and RLHF-based optimization~\citep{gunjal2024detecting, yu2024rlhf}. Another important direction is training-free inference-time intervention. Attentional intervention methods~\citep{SVAR_devils, PAI, huang2024opera} suppress hallucinations by modifying attention behaviors during decoding, but often incur extra inference overhead. In parallel, contrastive-decoding-based methods, such as SID~\citep{SID} and VCD~\citep{VCD}, steer the decoding distribution by contrasting different visual or textual conditions, though their effectiveness may be unstable due to the additional noise introduced in the contrastive process.
\section{Preliminary Study}\label{Preliminary Study}
\begin{figure*}[t]
    \centering
    \includegraphics[width=\textwidth]{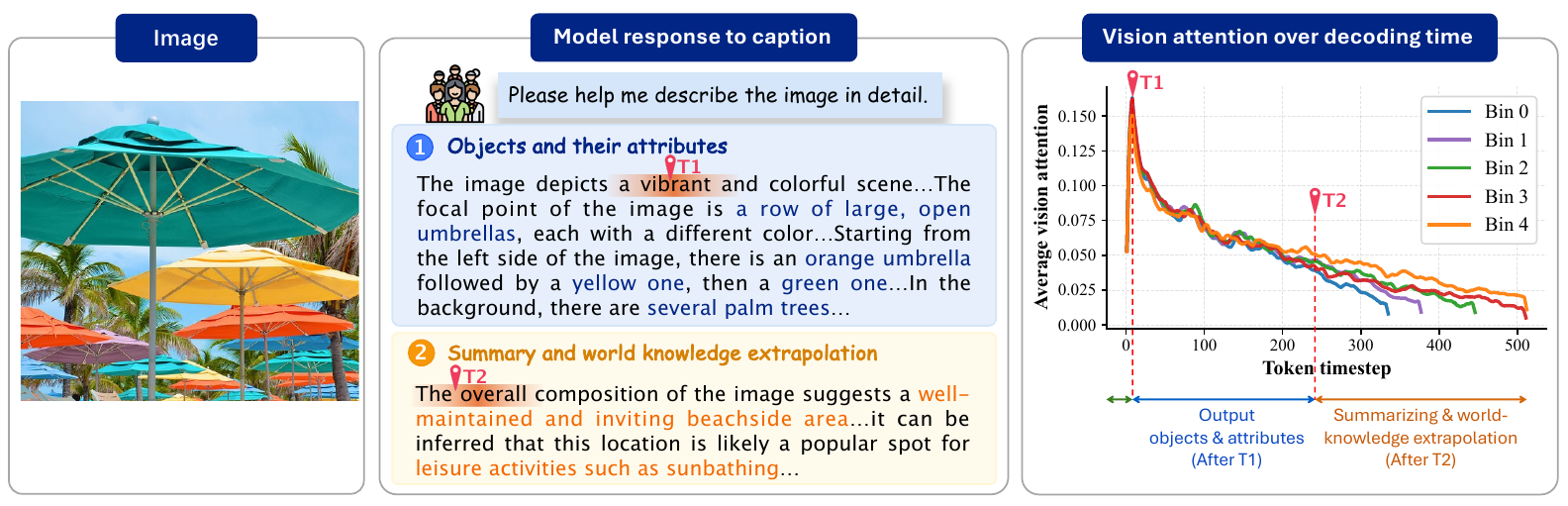}
    \caption{\textbf{Visual attention dynamics during decoding.}
    Results on 500 MSCOCO examples using Qwen2.5-VL-3B, grouped into five sequence-length bins. $T1$ and $T2$ mark the onset of visual description and semantic summarization, respectively.}
    \label{fig:timestep_examples}
\end{figure*}
We conduct preliminary experiments on the CHAIR benchmark using 500 MSCOCO examples with Qwen2.5-VL-3B. We examine the temporal dynamics of visual attention during generation, compare fixed and query-adaptive visual steering, and investigate the contributions of different context sources to hallucination mitigation. These analyses yield three key observations.
\begin{table}[h]
    \centering
    \caption{\textbf{Preliminary analysis of adaptive and multi-context steering.}
    (a) Comparison of fixed and adaptive query steering toward the vision prototype.
    (b) Impact of different context branches under fixed-strength steering. $C_S$ and $C_I$ denote CHAIR$_S$ and CHAIR$_I$, respectively.}
    \label{tab:preliminary}

    \fontsize{7}{9}\selectfont
    \renewcommand{\arraystretch}{1.1}

    \begin{minipage}[t]{0.49\linewidth}
        \centering
        \best{(a) Adaptive Visual Steering}

        \vspace{2pt}
        \setlength{\tabcolsep}{2.2pt}
        \begin{tabular}{
            l |
            c c
            >{\color{gray}}c
            >{\color{gray}}c
            >{\color{gray}}c
        }
            \Xhline{1pt}
            Method 
            & $C_{S}\downarrow$ 
            & $C_{I}\downarrow$
            & Recall$\uparrow$ 
            & F1$\uparrow$ 
            & Length$\uparrow$ \\
            \Xcline{1-6}{0.4pt}

            Baseline
            & 56.0 & 9.96 & 74.05 & 74.10 & 404.19 \\

            $\text{Fixed-V}_{\alpha=0.05}$
            & 51.6 & 9.48 & 74.11 & 74.99 & 350.88 \\

            $\text{Fixed-V}_{\alpha=0.15}$
            & \colorbox{gray!20}{\best{39.0}} & \underline{9.30} & 70.18 & 75.29 & \underline{199.70} \\

            $\text{Adaptive-V}$
            & \underline{40.2} & \colorbox{gray!20}{\best{8.33}}
            & 71.53 & 75.08 & 292.41 \\

            \Xhline{1pt}
        \end{tabular}
    \end{minipage}
    \hfill
    \begin{minipage}[t]{0.49\linewidth}
        \centering
        \best{(b) Multi-context Steering}

        \vspace{2pt}
        \setlength{\tabcolsep}{2.2pt}
        \begin{tabular}{
            l |
            c c
            >{\color{gray}}c
            >{\color{gray}}c
            >{\color{gray}}c
        }
            \Xhline{1pt}
            Method 
            & $C_{S}\downarrow$ 
            & $C_{I}\downarrow$
            & Recall$\uparrow$ 
            & F1$\uparrow$ 
            & Length$\uparrow$ \\
            \Xcline{1-6}{0.4pt}

            $\text{Fixed-V}$
            & 51.60 & 9.48 & 74.11 & 74.99 & 350.88 \\

            $\text{Fixed-V-G}$
            & 47.80 & 8.95 & 73.10 & 75.15 & 345.88 \\

            $\text{Fixed-P-G}$
            & \underline{47.62} & 9.35 & 73.54 & 74.68 & 382.81 \\

            $\text{Fixed-G}$
            & 50.61 & \colorbox{gray!20}{\best{8.80}} & 74.68 & 75.40 & 385.38 \\

            $\text{Fixed-V-P-G}$
            & \colorbox{gray!20}{\best{46.00}} & \underline{8.94}
            & 73.35 & 75.61 & 317.33 \\

            \Xhline{1pt}
        \end{tabular}
    \end{minipage}
\end{table}

\noindent\textbf{Observation 1. Vision attention naturally peaks when object grounding is required.}
To examine how visual reliance evolves throughout generation, we evenly divide the generated sequences into five length bins according to the number of samples and measure the average attention from the query to visual tokens at each generation timestep. 
As shown in Fig.~\ref{fig:timestep_examples}, despite substantial differences in sequence length, visual attention consistently peaks around $T1 \in \{4,5\}$, when generation moves beyond generic openings (e.g. “\textit{The image depicts a …}”) and begins grounding concrete objects and attributes. It then gradually declines as generation shifts toward summarization and world-knowledge extrapolation at $T2$, where less direct visual evidence is required. This pattern aligns with human intuition: during summarization and extrapolation, generation can rely on previously established context rather than direct visual evidence.

\noindent\textbf{Observation 2. Adaptive visual steering achieves a better hallucination-generation trade-off than fixed amplification.}
Table~\ref{tab:preliminary} (a) compares fixed visual steering with query–vision affinity-based adaptive steering. Strong fixed steering ($\alpha=0.15$) reduces $C_{S}$ to 39.0 but sharply shortens responses to 199.70 tokens. In contrast, $Adaptive$-$V$ achieves a comparable $C_{S}$ of 40.2 while retaining 292.41 tokens and reducing $C_{I}$ to 8.33, suggesting that visual evidence should be reinforced according to the model’s intrinsic vision-attending tendency rather than uniformly amplified.

\noindent\textbf{Observation 3. Hallucination mitigation benefits from multiple context branches beyond vision.}
As shown in Table~\ref{tab:preliminary} (b), textual contexts also provide useful corrective signals: $P+G$ reduces $C_{S}$ from 56.00 to 47.62 without visual reinforcement, while $G$ alone achieves a $C_{I}$ of 8.80. Combining $V$, $P$, and $G$ further yields the lowest $C_{S}$ of 46.00 among the branch combinations, motivating multi-source coordination rather than vision-only enhancement.
\section{method}\label{method}
\begin{figure*}[t]
    \centering
    \includegraphics[width=\textwidth]{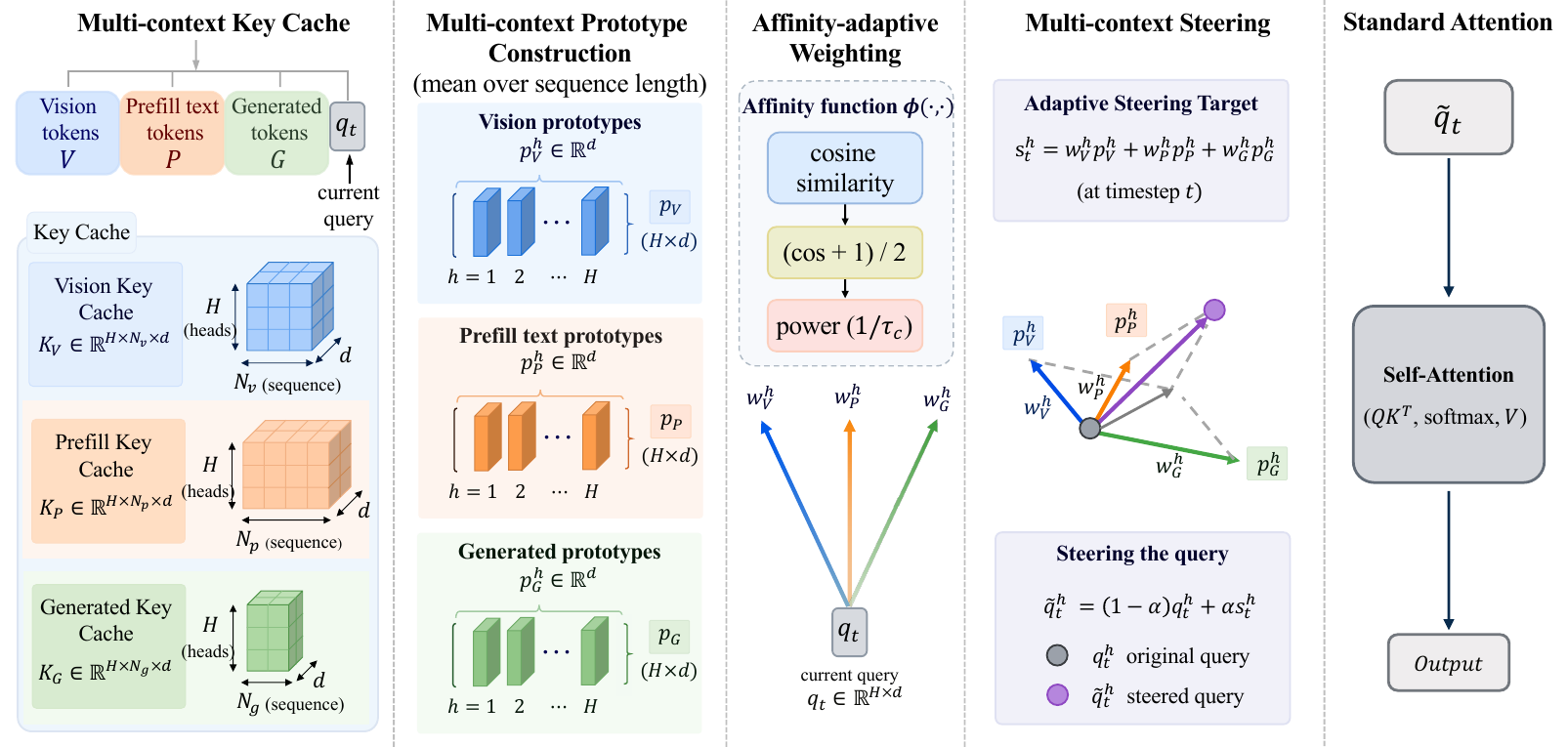}
    \caption{\textbf{Overview of AIMS.} Given vision, prefilled text, and generated-token key caches, AIMS constructs head-wise multi-context prototypes and adaptively weights them according to their affinity with the current query. The weighted prototypes form a dynamic steering target that adjusts the query representation before standard self-attention.}
    \label{fig:method main}
\end{figure*}

The above observations reveal a phenomenon overlooked by existing methods: LVLMs inherently exhibit a semantically meaningful vision-attending tendency aligned with their evolving generation needs, suggesting that hallucination mitigation can benefit from reinforcing, rather than overriding, this intrinsic tendency. Moreover, enhancing prefilled and generated contexts can also mitigate hallucinations, indicating that useful corrective signals extend beyond visual evidence. Motivated by these findings, we propose \textbf{AIMS} (\underline{A}daptive \underline{I}nformation \underline{M}ulti-source \underline{S}teering), a lightweight training-free framework that adaptively coordinates three context sources inherent to autoregressive generation: vision tokens ($V$) for perceptual evidence, prefilled text tokens ($P$) for task intent, and previously generated tokens ($G$) for contextual coherence. By estimating their query-dependent affinities, AIMS dynamically determines which contextual information should guide each decoding step, rather than relying on fixed or vision-centric steering.
\paragraph{Multi-context prototype construction.}
In Transformer attention, the relevance of contextual information to the current generation state is naturally captured through query--key matching. We therefore leverage the existing key cache to construct compact prototypes for each context source. Specifically, we identify vision tokens ($V$), prefilled text tokens ($P$), and previously generated tokens ($G$) by their token indices and mean-pool their key representations within each source. For attention head $h$, the prototype of context source $i$ is defined as
\begin{equation}
\mathbf{p}_{i}^{h}=
\frac{1}{N_i}
\sum_{j=1}^{N_i}
\mathbf{k}_{i,j}^{h},
\qquad
i\in\{V,P,G\},
\label{eq:prototype}
\end{equation}
where $\mathbf{k}_{i,j}^{h}$ denotes the key representation of the $j$-th token in context source $i$, and $N_i$ is the number of tokens in that source. The resulting prototypes $\mathbf{p}_V^h$, $\mathbf{p}_P^h$, and $\mathbf{p}_G^h$ provide compact summaries of the three context sources in the model's native query--key representation space, allowing their relevance to the current query to be measured in a unified manner. Since the prototypes are constructed directly from the existing key cache, this process requires no additional encoding or auxiliary forward passes and introduces only lightweight computational overhead.

\paragraph{Affinity-adaptive weighting.}
Observation~2 suggests that effective intervention should follow, rather than override, the model's intrinsic information-attending tendency. Since the current query encodes the generation state at timestep $t$, its compatibility with different context prototypes provides a natural internal signal of which information sources are currently relevant. We therefore use query--context affinity to adaptively determine the contribution of each context source.

Specifically, for context source $c\in\{V,P,G\}$ and attention head $h$, we measure the affinity between the current query $\mathbf{q}_t^h$ and the corresponding prototype $\mathbf{p}_{c,t}^h$ using cosine similarity and map it to $[0,1]$. The adaptive weight is defined as
\begin{equation}
w_{c,t}^h =
\left(
\frac{\cos(\mathbf{q}_t^h,\mathbf{p}_{c,t}^h)+1}{2}
\right)^{1/\tau_c},
\qquad c\in\{V,P,G\},
\label{eq:adaptive_weight}
\end{equation}
where $\tau_c$ controls the sensitivity of source $c$ to its query--context affinity. Consequently, the resulting weights are simultaneously source-, head-, and timestep-adaptive, allowing AIMS to capture fine-grained variations in contextual demand throughout autoregressive generation.

For notational simplicity, we use $\mathbf{p}_{c,t}^h$ to denote the prototype available at timestep $t$. The visual and prefilled contexts remain unchanged during decoding, and thus $\mathbf{p}_{V,t}^h=\mathbf{p}_V^h$ and $\mathbf{p}_{P,t}^h=\mathbf{p}_P^h$ are fixed. In contrast, the generated context grows autoregressively. Instead of repeatedly pooling an increasingly long generation history, we use the key representation of the latest generated token as $\mathbf{p}_{G,t}^h$. As this representation has already contextualized the preceding tokens through causal self-attention, it provides a naturally updated and lightweight summary of the generation history.

\paragraph{Multi-context steering.}
Rather than separately reweighting the attention logits of the $V$, $P$, and $G$ branches, AIMS first integrates their affinity-weighted prototypes into a unified steering target and uses it to update the current query. The updated query then interacts with all cached keys through the original attention operation, allowing the three context sources to jointly influence the attention assigned to individual tokens. Specifically, we construct a head-wise steering target by integrating the context prototypes according to their adaptive affinities:
\begin{equation}
\mathbf s_t^h
=
\sum_{c\in\{V,P,G\}}
w_{c,t}^{h}\mathbf p_{c,t}^{h}.
\label{eq:steering_direction}
\end{equation}
The current query is then updated as
\begin{equation}
\widetilde{\mathbf q}_t^h
=
(1-\alpha)\mathbf q_t^h
+
\alpha\mathbf s_t^h,
\label{eq:query_update}
\end{equation}
where $\alpha$ controls the overall steering strength. The modified query is subsequently passed to the original attention operation:
\begin{equation}
\mathbf{o^t_h}
=
\operatorname{Softmax}
\left(
\frac{
\widetilde{\mathbf{q}}^h_{t}
(\mathbf{K}^h)^{\top}
}{
\sqrt{d_h}
}
\right)
\mathbf{V^h}.
\end{equation}

AIMS modifies only the current query representation while leaving the key-value cache and the original attention computation unchanged, without requiring an additional model forward pass.

\section{experiment}
\begin{table}[h]
    \centering
    \caption{Evaluation results of different methods on CHAIR.}
	\label{tab:main_results}
    \fontsize{7}{9}\selectfont 
    \setlength{\tabcolsep}{3.2pt} 
	\renewcommand{\arraystretch}{1.1}
	\begin{tabular}{
        l |
        cc >{\color{gray}}c >{\color{gray}}c >{\color{gray}}c |
        cc >{\color{gray}}c >{\color{gray}}c >{\color{gray}}c |
        cc >{\color{gray}}c >{\color{gray}}c >{\color{gray}}c
    }
		\Xhline{1pt} 

		\multirow{3}{*}[-1pt]{\best{Method}} 
        & \multicolumn{15}{c}{\best{LLaVA-1.5-7B}} \\
        \Xcline{2-16}{0.4pt}

        & \multicolumn{5}{c|}{\best{Greedy}} 
        & \multicolumn{5}{c|}{\best{Beam Search}} 
        & \multicolumn{5}{c}{\best{Nucleus}} \\
        \Xcline{2-16}{0.4pt}

        & $C_S$$\downarrow$ 
        & $C_I$$\downarrow$ 
        & \textcolor{gray}{R$\uparrow$} 
        & \textcolor{gray}{F1$\uparrow$} 
        & \textcolor{gray}{Len$\uparrow$} 
        & $C_S$$\downarrow$ 
        & $C_I$$\downarrow$ 
        & \textcolor{gray}{R$\uparrow$} 
        & \textcolor{gray}{F1$\uparrow$} 
        & \textcolor{gray}{Len$\uparrow$} 
        & $C_S$$\downarrow$ 
        & $C_I$$\downarrow$ 
        & \textcolor{gray}{R$\uparrow$} 
        & \textcolor{gray}{F1$\uparrow$} 
        & \textcolor{gray}{Len$\uparrow$} \\
        \Xcline{1-16}{0.4pt}

		Baseline 
        & 51.4 & 14.70 & 79.42 & 76.80 & 97.9 
        & 55.2 & 14.93 & 80.34 & 77.10 & 103.5 
        & 50.0 & 15.31 & 74.05 & 73.19 & 101.5 \\

        MemVR 
        & 49.8 & 14.20 & 78.64 & 76.81 & 93.8 
        & 53.0 & 14.30 & 79.54 & 77.53 & 95.7 
        & 52.6 & 15.80 & 72.91 & 72.73 & 93.6 \\

        OPERA 
        & -- & -- & -- & -- & -- 
        & \underline{44.6} & \underline{12.72} & 79.25 & 78.40 & 92.7 
        & -- & -- & -- & -- & -- \\

        VCD 
        & -- & -- & -- & -- & -- 
        & -- & -- & -- & -- & -- 
        & 49.9 & 15.72 & 75.24 & 74.00 & 94.6 \\

        VTI 
        & 48.8 & \underline{13.89} & 78.43 & 77.06 & 94.1 
        & 48.8 & 12.87 & 79.57 & 78.74 & 97.9 
        & \underline{49.2} & \underline{14.98} & 73.86 & 73.85 & 95.8 \\

        SID 
        & \underline{47.6} & 14.43 & 76.93 & 76.76 & 94.7 
        & -- & -- & -- & -- & -- 
        & 52.0 & 15.51 & 75.38 & 74.60 & 95.5 \\

        AIMS 
        & \colorbox{gray!20}{\best{43.6}} & \colorbox{gray!20}{\best{13.17}} & 76.84 & 77.96 & 89.9 
        & \colorbox{gray!20}{\best{43.6}} & \colorbox{gray!20}{\best{12.02}} & 76.46 & 77.24 & 93.5 
        & \colorbox{gray!20}{\best{47.1}} & \colorbox{gray!20}{\best{15.23}} & 71.50 & 72.91 & 85.6 \\

		\Xcline{1-16}{1pt}

        \multirow{3}{*}[-1pt]{\best{Method}} 
        & \multicolumn{15}{c}{\best{Qwen2.5-VL-3B}} \\
        \Xcline{2-16}{0.4pt}

        & \multicolumn{5}{c|}{\best{Greedy}} 
        & \multicolumn{5}{c|}{\best{Beam Search}} 
        & \multicolumn{5}{c}{\best{Nucleus}} \\
        \Xcline{2-16}{0.4pt}

        & $C_S$$\downarrow$ 
        & $C_I$$\downarrow$ 
        & \textcolor{gray}{R$\uparrow$} 
        & \textcolor{gray}{F1$\uparrow$} 
        & \textcolor{gray}{Len$\uparrow$} 
        & $C_S$$\downarrow$ 
        & $C_I$$\downarrow$ 
        & \textcolor{gray}{R$\uparrow$} 
        & \textcolor{gray}{F1$\uparrow$} 
        & \textcolor{gray}{Len$\uparrow$} 
        & $C_S$$\downarrow$ 
        & $C_I$$\downarrow$ 
        & \textcolor{gray}{R$\uparrow$} 
        & \textcolor{gray}{F1$\uparrow$} 
        & \textcolor{gray}{Len$\uparrow$} \\
        \Xcline{1-16}{0.4pt}

		Baseline 
        & 56.0 & 9.96 & 74.05 & 74.10 & 404.2 
        & 45.1 & 8.93 & 73.41 & 76.19 & 367.3 
        & 56.6 & 11.39 & 72.96 & 72.50 & 441.3 \\

        MemVR 
        & 51.8 & 9.62 & 73.90 & 74.18 & 412.5 
        & 43.2 & 8.30 & 73.41 & 76.70 & 368.0 
        & 57.6 & 10.66 & 72.78 & 72.53 & 445.5 \\

        OPERA 
        & -- & -- & -- & -- & -- 
        & 45.2 & \underline{8.15} & 71.70 & 74.93 & 361.6 
        & -- & -- & -- & -- & -- \\

        VCD 
        & -- & -- & -- & -- & -- 
        & -- & -- & -- & -- & -- 
        & 57.8 & 9.99 & 73.86 & 73.00 & 477.0 \\

        VTI 
        & \underline{43.2} & \colorbox{gray!20}{\best{8.02}} & 70.88 & 75.02 & 338.2 
        & \underline{41.8} & 8.47 & 72.53 & 76.10 & 358.6 
        & 52.4 & \underline{9.78} & 71.51 & 73.00 & 430.0 \\

        SID 
        & 50.3 & 12.13 & 74.78 & 75.08 & 207.8 
        & -- & -- & -- & -- & -- 
        & \underline{50.6} & 10.15 & 73.35 & 74.25 & 386.4 \\

        AIMS 
        & \colorbox{gray!20}{\best{40.8}} & \underline{9.07} & 71.69 & 74.66 & 284.2 
        & \colorbox{gray!20}{\best{38.2}} & \colorbox{gray!20}{\best{7.67}} & 71.84 & 76.27 & 325.2 
        & \colorbox{gray!20}{\best{44.8}} & \colorbox{gray!20}{\best{9.33}} & 70.56 & 73.62 & 327.6 \\

		\Xcline{1-16}{1pt}

        \multirow{3}{*}[-1pt]{\best{Method}} 
        & \multicolumn{15}{c}{\best{Qwen3.5-9B}} \\
        \Xcline{2-16}{0.4pt}

        & \multicolumn{5}{c|}{\best{Greedy}} 
        & \multicolumn{5}{c|}{\best{Beam Search}} 
        & \multicolumn{5}{c}{\best{Nucleus}} \\
        \Xcline{2-16}{0.4pt}

        & $C_S$$\downarrow$ 
        & $C_I$$\downarrow$ 
        & \textcolor{gray}{R$\uparrow$} 
        & \textcolor{gray}{F1$\uparrow$} 
        & \textcolor{gray}{Len$\uparrow$} 
        & $C_S$$\downarrow$ 
        & $C_I$$\downarrow$ 
        & \textcolor{gray}{R$\uparrow$} 
        & \textcolor{gray}{F1$\uparrow$} 
        & \textcolor{gray}{Len$\uparrow$} 
        & $C_S$$\downarrow$ 
        & $C_I$$\downarrow$ 
        & \textcolor{gray}{R$\uparrow$} 
        & \textcolor{gray}{F1$\uparrow$} 
        & \textcolor{gray}{Len$\uparrow$} \\
        \Xcline{1-16}{0.4pt}

		Baseline 
        & \underline{47.4} & 11.07 & 74.11 & 75.87 & 224.3 
        & 50.0 & 10.82 & 75.19 & 76.08 & 234.0 
        & 51.0 & 12.03 & 73.86 & 74.74 & 227.1 \\

        MemVR 
        & 61.0 & 11.39 & 78.11 & 74.67 & 403.2 
        & 60.4 & 11.60 & 73.41 & 76.70 & 368.0 
        & 57.6 & 10.66 & 78.11 & 74.67 & 403.2 \\

        VCD 
        & -- & -- & -- & -- & -- 
        & -- & -- & -- & -- & -- 
        & 55.2 & 12.61 & 74.79 & 73.98 & 302.7 \\

        VTI 
        & 47.6 & \colorbox{gray!20}{\best{9.14}} & 69.23 & 72.98 & 355.6 
        & \underline{46.0} & \colorbox{gray!20}{\best{8.22}} & 68.59 & 72.99 & 356.8 
        & \underline{50.6} & \colorbox{gray!20}{\best{9.82}} & 68.78 & 71.95 & 357.9 \\

        SID 
        & 59.6 & 11.74 & 77.48 & 74.36 & 402.9 
        & -- & -- & -- & -- & -- 
        & 57.8 & 11.29 & 78.10 & 74.76 & 404.5 \\

        AIMS 
        & \colorbox{gray!20}{\best{40.0}} & \underline{9.91} & 67.05 & 73.38 & 249.9 
        & \colorbox{gray!20}{\best{42.6}} & \underline{9.45} & 69.52 & 74.52 & 270.9 
        & \colorbox{gray!20}{\best{47.0}} & \underline{10.62} & 69.56 & 72.57 & 256.8 \\

		\Xcline{1-16}{1pt}
	\end{tabular}
\end{table}
\subsection{EXPERIMENTAL SETTINGS}

\textbf{Models and Baselines.}
We evaluate AIMS on three representative LVLMs, including LLaVA-1.5-7B~\citep{llava1.5}, Qwen2.5-VL-3B~\citep{qwen2.5vl}, and Qwen3.5-9B. 
Since AIMS is a training-free framework, we compare it with representative hallucination mitigation methods, including VCD~\citep{VCD}, SID~\citep{SID}, VTI~\citep{VTI}, OPERA~\citep{huang2024opera}, MemVR~\citep{memvr} and Baseline. 

For decoding strategies, we evaluate AIMS under three generation settings: greedy decoding, beam search with 5 beams, and nucleus sampling with temperature=1, top-p=0.9, and top-k=50. For comparison methods, we follow the decoding strategies supported in their original papers.

\textbf{Datasets.}
\noindent\textbf{CHAIR}~\citep{chair} evaluates object hallucination in image captioning. Following previous works, we randomly sample 500 MSCOCO~\citep{coco2014} images and report $\mathrm{CHAIR}_I$ ($C_I$), $\mathrm{CHAIR}_S$ ($C_S$), Recall ($R$), F1, and generation length ($Len$).
\noindent\textbf{AMBER-G}~\citep{wang2023amber} evaluates object, attribute, and relation hallucinations in free-form descriptions. We follow the official protocol and report $CHAIR$, $Cover$, $Hal$, and $Cog$.
\noindent\textbf{FaithScore}~\citep{jing2024faithscore} evaluates free-form hallucinations through atomic-fact verification. We evaluate on LLaVA-1k and report FaithScore and Sentence-FaithScore.
\noindent\textbf{MME}~\citep{mme_bench} evaluates perception and cognition across 14 subtasks. We report perception ($Per$), cognition ($Cog$), and total ($Sum$) scores.

\begin{table}[h]
    \centering
    \caption{Evaluation results of different methods on AMBER-G.}
	\label{tab:main_results_amber_g}
        \fontsize{7}{9}\selectfont
        \setlength{\tabcolsep}{3.2pt} 
	\renewcommand{\arraystretch}{1.1}
	\begin{tabular}{l | cccc | cccc | cccc}
        \Xhline{1pt} 
		\multirow{3}{*}[-1pt]{\best{Method}} & \multicolumn{12}{c}{\best{Qwen2.5-VL-3B}} \\
        \Xcline{2-13}{0.4pt}
        & \multicolumn{4}{c|}{\best{Greedy}} & \multicolumn{4}{c|}{\best{Beam Search}} & \multicolumn{4}{c}{\best{Nucleus}} \\ 
        \Xcline{2-13}{0.4pt}
        & CHAIR$\downarrow$ & Cover$\uparrow$ & Hal$\downarrow$ & Cog$\downarrow$ & CHAIR$\downarrow$ & Cover$\uparrow$ & Hal$\downarrow$ & Cog$\downarrow$ & CHAIR$\downarrow$ & Cover$\uparrow$ & Hal$\downarrow$ & Cog$\downarrow$ \\
        \Xcline{1-13}{0.4pt}
		Baseline & 8.2 & 69.6 & 52.7 & 5.7 & 6.8 & 68.3 & 44.7 & 4.9 & 9.9 & 70.3 & 63.7 & 7.4 \\
        VTI  & 6.9 & \best{65.0} & 43.8 & 4.3 & 5.8 & 63.8 & 35.1 & 3.9 & 8.5 & \best{65.9} & 51.6 & 5.4 \\
        AIMS  & \best{6.6} & 64.8 & \best{39.7} & \best{3.8} & \best{5.2} & \best{64.0} & \best{33.2} & \best{3.1} & \best{7.8} & 65.2 & \best{47.9} & \best{4.2} \\

		\Xhline{1pt} 
		\multirow{3}{*}[-1pt]{\best{Method}} & \multicolumn{12}{c}{\best{Qwen3.5-9B}} \\
        \Xcline{2-13}{0.4pt}
        & \multicolumn{4}{c|}{\best{Greedy}} & \multicolumn{4}{c|}{\best{Beam Search}} & \multicolumn{4}{c}{\best{Nucleus}} \\ 
        \Xcline{2-13}{0.4pt}
        & CHAIR$\downarrow$ & Cover$\uparrow$ & Hal$\downarrow$ & Cog$\downarrow$ & CHAIR$\downarrow$ & Cover$\uparrow$ & Hal$\downarrow$ & Cog$\downarrow$ & CHAIR$\downarrow$ & Cover$\uparrow$ & Hal$\downarrow$ & Cog$\downarrow$ \\
        \Xcline{1-13}{0.4pt}
		Baseline & 10.4 & 75.7 & 71.8 & 5.5 & 9.6 & 76.4 & 69.3 & 5.3 & 10.2 & 75.6 & 70.5 & 5.0 \\
        VTI  & 9.6 & 75.2 & 69.1 & \best{5.4} & 9.3 & 75.6 & 69.4 & \best{5.3} & 10.0 & \best{75.4} & 71.8 & 5.8 \\
        AIMS  & \best{9.4} & \best{75.3} & \best{68.9} & 5.7 & \best{9.1} & \best{75.9} & \best{65.8} & 5.6 & \best{9.4} & 74.8 & \best{65.9} & \best{5.4}\\
		\Xcline{1-13}{1pt}
	\end{tabular}
\end{table}

\noindent\textbf{Implementation details.}
We mainly consider $(\tau_V,\tau_P,\tau_G)\in\{(1.0,1.0,1.0),  (1.5,1.2,1.1)\}$ for the domain-specific sensitivity parameters in Eq.~\ref{eq:adaptive_weight}. The steering strength $\alpha$ in Eq.~\ref{eq:query_update} is set to $0.05$ or $0.08$ for LLaVA-1.5 and Qwen2.5-VL, and to $0.3$ or $0.4$ for Qwen3.5. All experiments are conducted on NVIDIA GPUs for consistent hardware evaluation.
\begin{figure*}[t]
    \centering

    \begin{minipage}[t]{0.5\textwidth}
    \vspace{0pt}
        \centering
        \captionof{table}{
            FaithScore evaluation results on Qwen2.5-VL-3B.
            FS and S-FS denote FaithScore and sentence-level
            FaithScore, respectively.
        }
        \label{tab:main_results_faithscore}

        \vspace{2pt}
        \fontsize{7}{9}\selectfont
        \setlength{\tabcolsep}{4pt}
        \renewcommand{\arraystretch}{1.1}

        \begin{tabular}{l c | cc}
            \Xhline{1pt}
            \best{Method} & \best{Decoding}
            & FS$\uparrow$
            & S-FS$\uparrow$ \\
            \Xhline{0.4pt}

            Baseline & \multirow{3}{*}{Greedy}
                     & 0.944 & 0.786 \\
            VTI      & & 0.942 & 0.787 \\
            AIMS     & & \best{0.945} & \best{0.791} \\

            \Xcline{1-4}{0.4pt}

            Baseline & \multirow{3}{*}{Beam}
                     & 0.942 & 0.780 \\
            VTI      & & 0.947 & 0.793 \\
            AIMS     & & \best{0.948} & \best{0.854} \\

            \Xcline{1-4}{0.4pt}

            Baseline & \multirow{3}{*}{Nucleus}
                     & 0.925 & 0.745 \\
            VTI      & & 0.927 & 0.746 \\
            AIMS     & & \best{0.929} & \best{0.789} \\

            \Xhline{1pt}
        \end{tabular}
    \end{minipage}
    \hfill
    \begin{minipage}[t]{0.45\textwidth}
    \vspace{0pt}
        \centering
        \includegraphics[width=\linewidth]{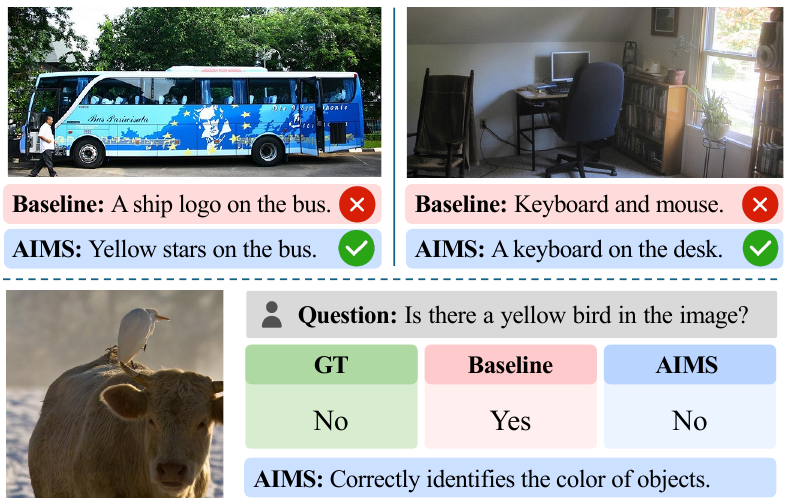}
        \captionof{figure}{Qualitative comparison between the baseline and AIMS.}
        \label{fig:case_study_zhengwen}
    \end{minipage}
\end{figure*}
\begin{table}[h]
    \centering
    \caption{Evaluation results of different methods on MME.}
	\label{tab:main_results_mme}
        \fontsize{7}{9}\selectfont 
        \setlength{\tabcolsep}{3.2pt} 
	\renewcommand{\arraystretch}{1.1}
	\begin{tabular}{l | ccc | ccc | ccc}
		\Xhline{1pt} 

        %
		\multirow{3}{*}[-1pt]{\best{Method}} & \multicolumn{9}{c}{\best{Qwen2.5-VL-3B}} \\
        \Xcline{2-10}{0.4pt}
        & \multicolumn{3}{c|}{\best{Greedy}} & \multicolumn{3}{c|}{\best{Beam Search}} & \multicolumn{3}{c}{\best{Nucleus}} \\ 
        \Xcline{2-10}{0.4pt}
        & Per$\uparrow$ & Cog$\uparrow$ & Sum$\uparrow$ & Per$\uparrow$ & Cog$\uparrow$ & Sum$\uparrow$ & Per$\uparrow$ & Cog$\uparrow$ & Sum$\uparrow$ \\
        \Xcline{1-10}{0.4pt}
		Baseline & 1591.38 & 612.86 & 2204.24 & 1558.17 & 537.14 & 2095.31 & 1466.52 & 513.93 & 1980.45 \\
        MemVR & 1591.93 & 626.42 & \underline{2218.36} & 1538.66 & 544.64 & \underline{2083.30} & 1425.96 & 523.92 & 1949.88 \\
        VCD & -- & -- & -- & -- & -- & -- & 1485.20 & 521.78 & \underline{2006.98} \\
        VTI & 1472.86 & 522.50 & 1995.36 & 1520.20 & 480.35 & 2000.55 & 1448.89 & 518.57 & 1967.46 \\
        SID & 1590.13 & 606.78 & 2196.91 & -- & -- & -- & 1450.47 & 557.14 & \colorbox{gray!20}{\best{2007.61}} \\
        AIMS & 1591.23 & 630.35 & \colorbox{gray!20}{\best{2221.58}} & 1550.71 & 540.00 & \colorbox{gray!20}{\best{2090.71}} & 1492.41 & 487.82 & 1980.23 \\
        
		\Xhline{1pt} 
		\multirow{3}{*}[-1pt]{\best{Method}} & \multicolumn{9}{c}{\best{Qwen3.5-9B}} \\
        \Xcline{2-10}{0.4pt}
        & \multicolumn{3}{c|}{\best{Greedy}} & \multicolumn{3}{c|}{\best{Beam Search}} & \multicolumn{3}{c}{\best{Nucleus}} \\ 
        \Xcline{2-10}{0.4pt}
        & Per$\uparrow$ & Cog$\uparrow$ & Sum$\uparrow$ & Per$\uparrow$ & Cog$\uparrow$ & Sum$\uparrow$ & Per$\uparrow$ & Cog$\uparrow$ & Sum$\uparrow$ \\
        \Xcline{1-10}{0.4pt}
		Baseline & 1714.86 & 678.57 & 2393.43 & 1722.59 & 584.64 & 2307.23 & 1587.53 & 567.85 & 2155.38 \\
        MemVR & 1707.45 & 671.07 & \colorbox{gray!20}{\best{2378.52}} & 1709.63 & 574.72 & \underline{2284.35} & 1580.36 & 569.08 & 2149.44 \\
        VCD & -- & -- & -- & -- & -- & -- & 1603.84 & 589.21 & 2193.05 \\
        VTI & 1631.79 & 592.14 & 2223.93 & 1652.71 & 508.21 & 2160.92 & 1501.74 & 535.00 & 2036.74 \\
        SID & 1693.31 & 656.07 & 2349.38 & -- & -- & -- & 1620.75 & 593.92 & \underline{2214.67} \\
        AIMS & 1699.97 & 653.92 & \underline{2353.90} & 1710.11 & 598.21 & \colorbox{gray!20}{\best{2308.32}} & 1625.41 & 591.78 & \colorbox{gray!20}{\best{2217.20}} \\
		\Xcline{1-10}{1pt}
	\end{tabular}
\end{table}

\subsection{EXPERIMENTAL RESULTS}
\paragraph{Results on CHAIR.}
As shown in Table~\ref{tab:main_results}, AIMS achieves the lowest $C_S$ in all nine model-decoding settings, consistently outperforming the baseline and existing training-free methods. Notably, AIMS maintains a consistent advantage over VTI across all three LVLMs and decoding strategies. While VTI applies pre-computed intervention directions consistently across queries, AIMS adapts its steering to the current generation state. This query-dependent coordination may explain its stable advantage across different decoding dynamics. Compared with contrastive decoding methods such as VCD and SID, AIMS also achieves stronger hallucination reduction without constructing additional contrastive inputs or auxiliary forward passes.

\paragraph{Results on AMBER-G.}
As shown in Table~\ref{tab:main_results_amber_g}, AIMS consistently achieves lower $Hal$ than VTI across all six model-decoding settings, indicating fewer hallucinated responses, while maintaining comparable $Cover$ scores and thus similar object coverage. This suggests that its hallucination reduction does not simply result from mentioning fewer objects present in the image. Beyond this general trend, AIMS also reduces $Cog$ across all decoding strategies on Qwen2.5-VL-3B, indicating a lower tendency to generate the human-cognition-associated hallucinated objects captured by AMBER. Together with the consistently lower $CHAIR$ scores, these results show that AIMS improves hallucination control at both the object and response levels without a substantial coverage trade-off.
\paragraph{Results on FaithScore.}
As shown in Table~\ref{tab:main_results_faithscore}, AIMS achieves the highest FS and S-FS across all three decoding strategies, with particularly pronounced sentence-level gains under beam search and nucleus sampling. Unlike object-centric metrics, FaithScore evaluates descriptive content by decomposing it into atomic facts and verifying their consistency with the image. The improvements therefore provide complementary evidence that AIMS extends beyond reducing hallucinated objects to improving the faithfulness of fine-grained visual facts in free-form responses.
\paragraph{Results on General-purpose Benchmark.}
Together with Tables~\ref{tab:main_results}–\ref{tab:main_results_faithscore}, Table~\ref{tab:main_results_mme} shows that AIMS largely preserves the general-purpose multimodal capabilities of the original models while substantially reducing hallucinations. VTI achieves effective hallucination mitigation but exhibits noticeable MME degradation on the two Qwen models, whereas MemVR better preserves general capabilities but provides more limited hallucination reduction. Overall, AIMS achieves a better balance between hallucination mitigation and general-purpose capability preservation.
\begin{table}[h]
    \centering
    \caption{Ablation of different affinity weighting functions on Qwen2.5-VL-3B.}
	\label{tab:aaw_ablation}
    \fontsize{7}{9}\selectfont 
    \setlength{\tabcolsep}{3.2pt} 
	\renewcommand{\arraystretch}{1.1}
	\begin{tabular}{
        l |
        cc >{\color{gray}}c >{\color{gray}}c >{\color{gray}}c |
        cc >{\color{gray}}c >{\color{gray}}c >{\color{gray}}c |
        cc >{\color{gray}}c >{\color{gray}}c >{\color{gray}}c
    }
		\Xhline{1pt} 

		\multirow{2}{*}[-1pt]{\best{Method}} 
        & \multicolumn{5}{c|}{\best{Greedy}} 
        & \multicolumn{5}{c|}{\best{Beam Search}} 
        & \multicolumn{5}{c}{\best{Nucleus}} \\
        \Xcline{2-16}{0.4pt}

        & $C_S$$\downarrow$ 
        & $C_I$$\downarrow$ 
        & \textcolor{gray}{R$\uparrow$} 
        & \textcolor{gray}{F1$\uparrow$} 
        & \textcolor{gray}{Len$\uparrow$} 
        & $C_S$$\downarrow$ 
        & $C_I$$\downarrow$ 
        & \textcolor{gray}{R$\uparrow$} 
        & \textcolor{gray}{F1$\uparrow$} 
        & \textcolor{gray}{Len$\uparrow$} 
        & $C_S$$\downarrow$ 
        & $C_I$$\downarrow$ 
        & \textcolor{gray}{R$\uparrow$} 
        & \textcolor{gray}{F1$\uparrow$} 
        & \textcolor{gray}{Len$\uparrow$} \\
        \Xcline{1-16}{0.4pt}


        AIMS$_{RBF}$ 
        & 42.8 & 8.85 & 71.51 & 75.31 & 301.5 
        & 38.4 & 7.87 & 73.67 & 77.40 & 311.2 
        & 51.0 & 9.74 & 70.56 & 72.94 & 339.9 \\

        AIMS$_{NCS}$ 
        & \colorbox{gray!20}{\best{40.2}} 
        & \colorbox{gray!20}{\best{8.33}} 
        & 71.53 & 75.08 & 292.4 
        & 38.4 & 7.84 & 72.84 & 77.02 & 325.9 
        & \underline{45.6} & 9.64 & 70.75 & 73.74 & 335.3 \\

        AIMS$_{KL}$ 
        & 46.9 & \underline{8.48} & 71.40 & 74.27 & 344.7 
        & \colorbox{gray!20}{\best{36.2}} 
        & \colorbox{gray!20}{\best{7.57}} 
        & 72.84 & 77.41 & 314.7 
        & 51.8 & \underline{9.43} & 71.83 & 72.54 & 399.6 \\

        AIMS 
        & \underline{40.8} & 9.07 & 71.69 & 74.66 & 284.2 
        & \underline{38.2} & \underline{7.67} & 71.84 & 76.27 & 325.2 
        & \colorbox{gray!20}{\best{44.8}} 
        & \colorbox{gray!20}{\best{9.33}} 
        & 70.56 & 73.62 & 327.6 \\

		\Xcline{1-16}{1pt}
	\end{tabular}
\end{table}
\subsection{ABLATION AND ANALYSIS}
\noindent\textbf{Ablation on steering configurations.}
We further examine the steering configurations of AIMS on Qwen2.5-VL-3B using CHAIR under greedy decoding, including the choice of context branches, adaptive weighting, and key hyperparameters. For qualitative illustration, Figure~\ref{fig:case_study_zhengwen} presents several examples, with additional case studies provided in Appendix~\ref{sec:case_study}.
\begin{wrapfigure}{r}{0.4\columnwidth}
    \centering
    \includegraphics[width=0.4\columnwidth]{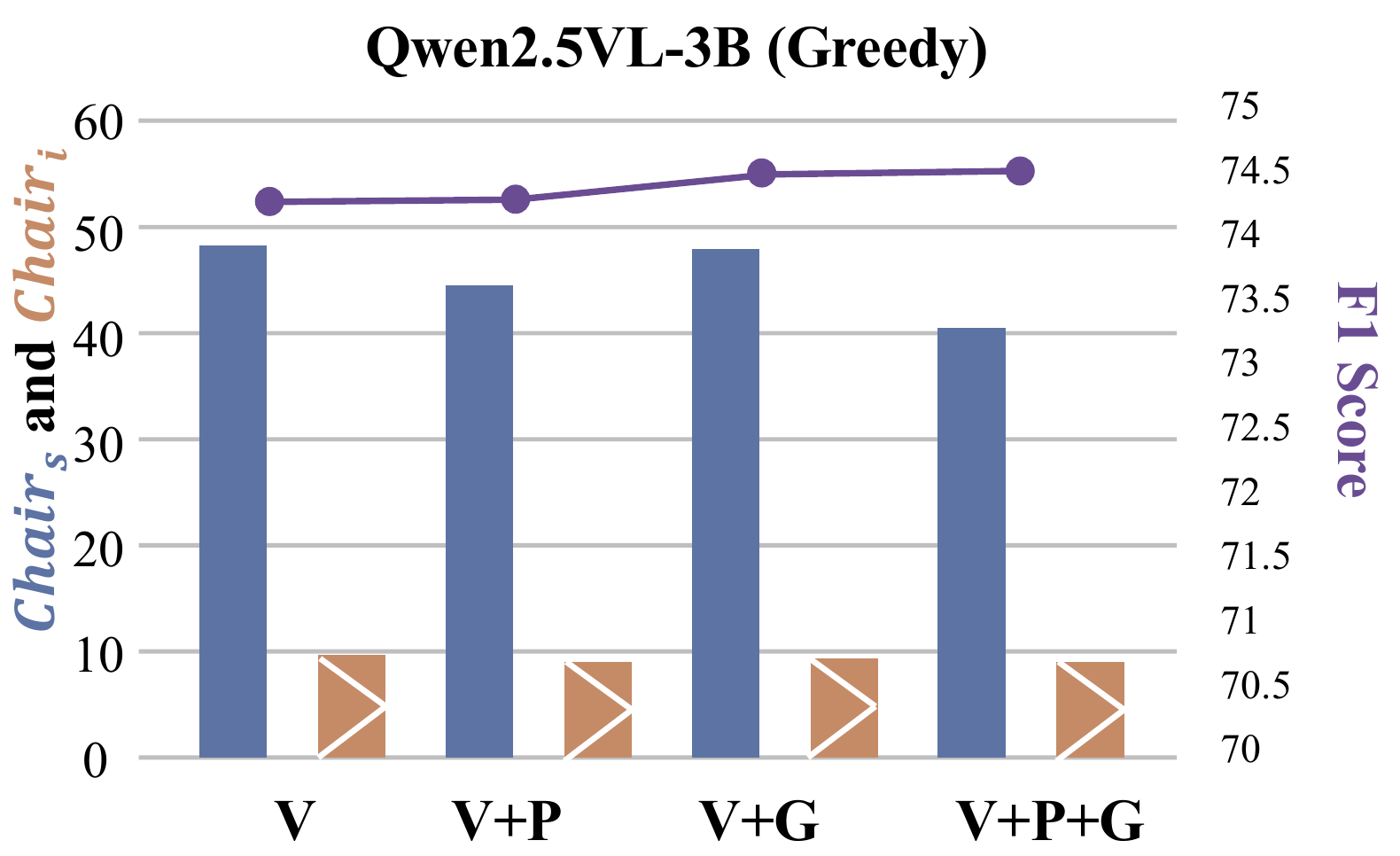}
    \caption{Ablation study of different context branches in AIMS.}
    \label{fig:3branchs_ablation}
\end{wrapfigure}

\textbf{Context Branches. }
Figure~\ref{fig:3branchs_ablation} investigates the contributions of
context branches in AIMS. Combining the visual branch with
either prefilled text context ($V+P$) or generation history
($V+G$) exhibits trade-offs across the evaluation metrics.
In contrast, jointly incorporating all three branches ($V+P+G$)
achieves the best performance on $CHAIR$ and $F1$. This demonstrates
the complementary roles of different contextual sources and supports
the use of multi-source steering in AIMS.

\noindent\textbf{Ablation on adaptive weighting functions.}
As shown in Table~\ref{tab:aaw_ablation}, AIMS reduces hallucination with different affinity functions, showing its robustness to the weighting function (see Appendix~\ref{sec:steering_weights} for function details).

\begin{figure*}[t]
    \centering
    \begin{minipage}[c]{0.68\textwidth}
        \centering
        \includegraphics[width=\linewidth]{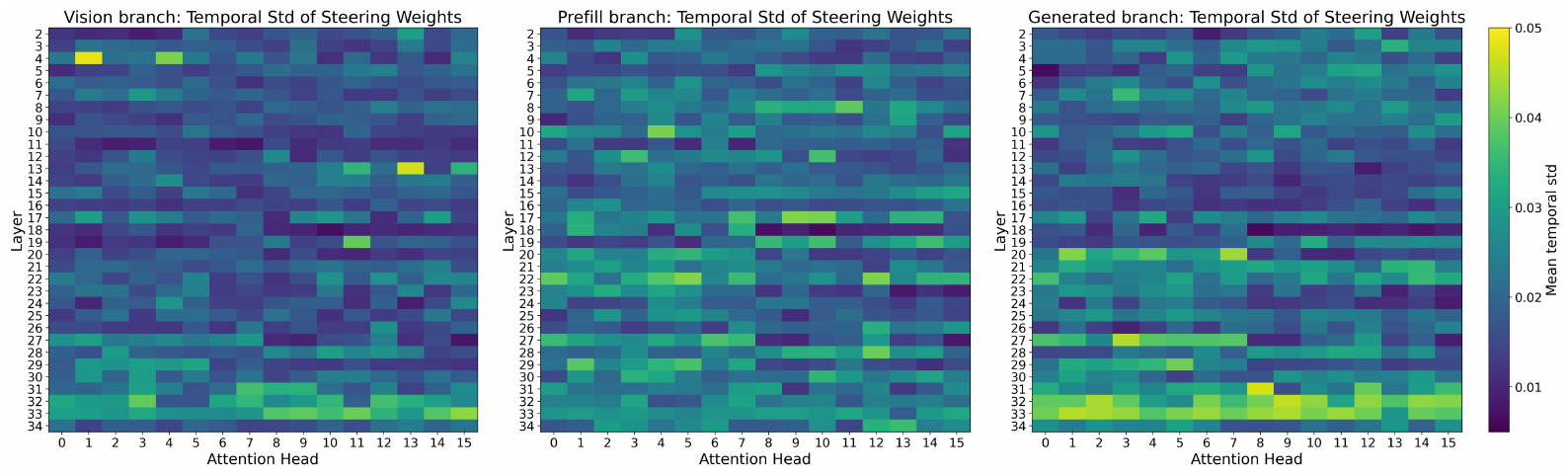}
        \captionof{figure}{
            Temporal variability of affinity-adaptive weights across different contextual branches. From left to right: $V$, $P$, and $G$.
        }
        \label{fig:vpg_timestep_heatmap}
    \end{minipage}
    \hfill
    \begin{minipage}[h]{0.3\textwidth}
        \centering
        \captionof{table}{Fixed vs. Affinity-adaptive weighting.}
        \label{tab:aaw}

        \fontsize{7}{9}\selectfont
        \setlength{\tabcolsep}{3.2pt}
        \renewcommand{\arraystretch}{1.1}
        \resizebox{\linewidth}{!}{%
        \begin{tabular}{l | ccc}
            \Xhline{1pt}
            \best{Method} 
            & $C_S$ $\downarrow$ 
            & $C_I$ $\downarrow$ 
            & F1 $\uparrow$ \\
            \Xhline{0.4pt}
            Baseline & 56.0 & 9.96 & 74.10 \\
            Fixed ($\alpha=0.1$) & 8.0 & 4.64 & 56.90 \\
            Fixed ($\alpha=0.01$) & 54.0 & 10.11 & 74.45 \\
            AIMS ($\alpha=0.08$) & 40.8 & 9.07 & 74.66 \\
            \Xhline{1pt}
        \end{tabular}%
        }
    \end{minipage}
\end{figure*}

\textbf{Affinity-adaptive Weighting. }
Table~\ref{tab:aaw} compares fixed-strength steering with the affinity-adaptive weighting used in AIMS. 
While strong fixed steering ($\alpha=0.1$) substantially reduces hallucination, it severely degrades $F1$, indicating excessive intervention in generation. 
More importantly, under comparable $F1$ scores (the last two rows), AIMS achieves substantially lower $C_S$ and $C_I$, demonstrating a better trade-off between hallucination mitigation and response quality.

\textbf{$\alpha$ and $\tau$. }
Table~\ref{tab:alpha tau} further studies the sensitivity to the steering strength $\alpha$ and the branch-specific temperatures $\tau$, where a larger $\tau$ indicates a broader sensitivity bandwidth to the corresponding context branch. Across different configurations of both $\alpha$ and $\tau$, AIMS consistently reduces hallucination compared with the original baseline while maintaining competitive $F1$ scores, demonstrating its robustness to these hyperparameter choices.

\begin{wraptable}{l}{0.48\columnwidth}
    \centering
    \caption{Ablation of $\alpha$ and $\tau$.}
    \label{tab:alpha tau}

    \fontsize{7}{9}\selectfont
    \setlength{\tabcolsep}{2.5pt}
    \renewcommand{\arraystretch}{1.1}
    \begin{tabular}{l | cc | ccc}
        \Xhline{1pt}
        \best{Method}
        & $\boldsymbol{\alpha}$
        & $(\boldsymbol{\tau_V},\boldsymbol{\tau_P},\boldsymbol{\tau_G})$
        & $C_S$ $\downarrow$
        & $C_I$ $\downarrow$
        & F1 $\uparrow$ \\
        \Xhline{0.4pt}
        \multirow{3}{*}{AIMS}
        & 0.08 & (1.5, 1.2, 1.1) & 40.8 & 9.07 & 74.66 \\
        & 0.05 & (1.5, 1.2, 1.1) & 49.8 & 8.95 & 75.51 \\
        & 0.08 & (1.0, 1.0, 1.0) & 44.4 & 7.59 & 75.22 \\
        \Xhline{1pt}
    \end{tabular}
\end{wraptable}
\textbf{Overhead Analysis.}
Table~\ref{tab:overhead} reports the overhead measured on a fixed subset of 10 CHAIR images. 
AIMS introduces nearly no additional peak memory over the baseline and achieves the second-lowest inference latency, closely approaching MemVR, demonstrating a favorable trade-off between effectiveness and computational efficiency.

\begin{wraptable}{r}{0.46\columnwidth}
    \centering
    \caption{Comparison of computational overhead on Qwen2.5-VL-3B.}
    \label{tab:overhead}

    \fontsize{7}{9}\selectfont
    \setlength{\tabcolsep}{3.2pt}
    \renewcommand{\arraystretch}{1.1}
    \begin{tabular}{l | cc}
        \Xhline{1pt}
        \best{Method}
        & Peak Mem. (MB) $\downarrow$
        & Time (s/token) $\downarrow$ \\
        \Xhline{0.4pt}
        Baseline & 7235.87 & 0.0437 \\
        MemVR & \underline{7239.53} & \best{0.0698} \\
        VCD & 7387.08 & 0.0905 \\
        VTI & 7482.31 & 0.2424 \\
        SID & 7759.74 & 0.1075 \\
        AIMS & \best{7236.11} & \underline{0.0705} \\
        \Xhline{1pt}
    \end{tabular}
\end{wraptable}
\textbf{Temporal Adaptivity of Steering Weights.}
We measure the temporal variability of steering weights using their standard deviation across decoding timesteps. As shown in Figure~\ref{fig:vpg_timestep_heatmap}, the three branches exhibit distinct layer- and head-specific patterns. For $V$, large variations are sparsely concentrated in individual heads at earlier layers (e.g., Layers 4, 13, and 19), but become broadly distributed across heads in Layers 31--33. $P$ shows more dispersed variations across middle and deep layers, whereas $G$ exhibits stronger layer-wise structures around Layers 20, 27, and 31--33. These distinct patterns suggest different dynamic utilization of contextual sources during decoding, motivating fine-grained affinity-adaptive multi-source steering across timesteps.

\section{Conclusion}
In this work, we revisit object hallucination in LVLMs from the perspective of multi-context interaction. Our analysis systematically reveals an intrinsic vision-attending tendency during generation and shows that complementary contextual information beyond visual evidence can also contribute to hallucination mitigation. Motivated by these findings, we propose AIMS, a training-free method that constructs compact context prototypes and dynamically coordinates their contributions during decoding. Extensive experiments across different LVLM architectures and decoding strategies demonstrate that AIMS consistently reduces object hallucination while preserving strong general-purpose multimodal capabilities. We hope our findings encourage a shift from visual-centric intervention toward adaptive multi-context coordination for hallucination mitigation.


\bibliographystyle{plainnat}
\bibliography{references}

\clearpage
\beginappendix
\section{Appendix}
\subsection{Alternative Affinity Weighting Functions}\label{sec:steering_weights}
We evaluate three alternative affinity weighting functions in addition to our default formulation, with results reported in Table~\ref{tab:aaw_ablation}. Here, $\mathbf{q}_t^h$, $\mathbf{p}_{c,t}^h$, and $\tau_c$ follow the notation in Sec.~\ref{method}.
\paragraph{RBF.}
The RBF-based weighting measures the Euclidean distance between the query and contextual prototype, assigning larger weights to closer representations:
\begin{equation}
w_{c,t}^h =
\exp\left(
-\frac{
\left\|\mathbf{q}_t^h-\mathbf{p}_{c,t}^h\right\|_2^2
}{
2\tau_c^2 d_h
}
\right),
\qquad c\in\{V,P,G\},
\label{eq:weight_rbf}
\end{equation}
where $d_h$ denotes the dimensionality of each attention head.
\paragraph{NCS.}
The nonlinear cosine similarity (NCS) first computes the normalized inner product between the query and contextual prototype, followed by softplus and sigmoid transformations:
\begin{equation}
w_{c,t}^h =
\operatorname{sigmoid}\left(
\operatorname{softplus}\left(
\frac{
\cos(\mathbf{q}_t^h,\mathbf{p}_{c,t}^h)
}{
\tau_c
}
\right)
\right),
\qquad c\in\{V,P,G\}.
\label{eq:weight_ncs}
\end{equation}
\paragraph{KL.}
For KL-based weighting, we first transform the query and contextual prototype into probability distributions:
\begin{equation}
\boldsymbol{\pi}_{q,t}^h =
\operatorname{softmax}\left(
\frac{\mathbf{q}_t^h}{\tau_c}
\right),
\qquad
\boldsymbol{\pi}_{c,t}^h =
\operatorname{softmax}\left(
\frac{\mathbf{p}_{c,t}^h}{\tau_c}
\right).
\end{equation}
The steering weight is then determined by their KL divergence:
\begin{equation}
w_{c,t}^h =
\exp\left(
-D_{\mathrm{KL}}
\left(
\boldsymbol{\pi}_{q,t}^h
\,\|\, 
\boldsymbol{\pi}_{c,t}^h
\right)
\right),
\qquad c\in\{V,P,G\}.
\label{eq:weight_kl}
\end{equation}
\subsection{Case Studies}\label{sec:case_study}
\begin{figure*}[t]
    \centering
    \includegraphics[width=\textwidth]{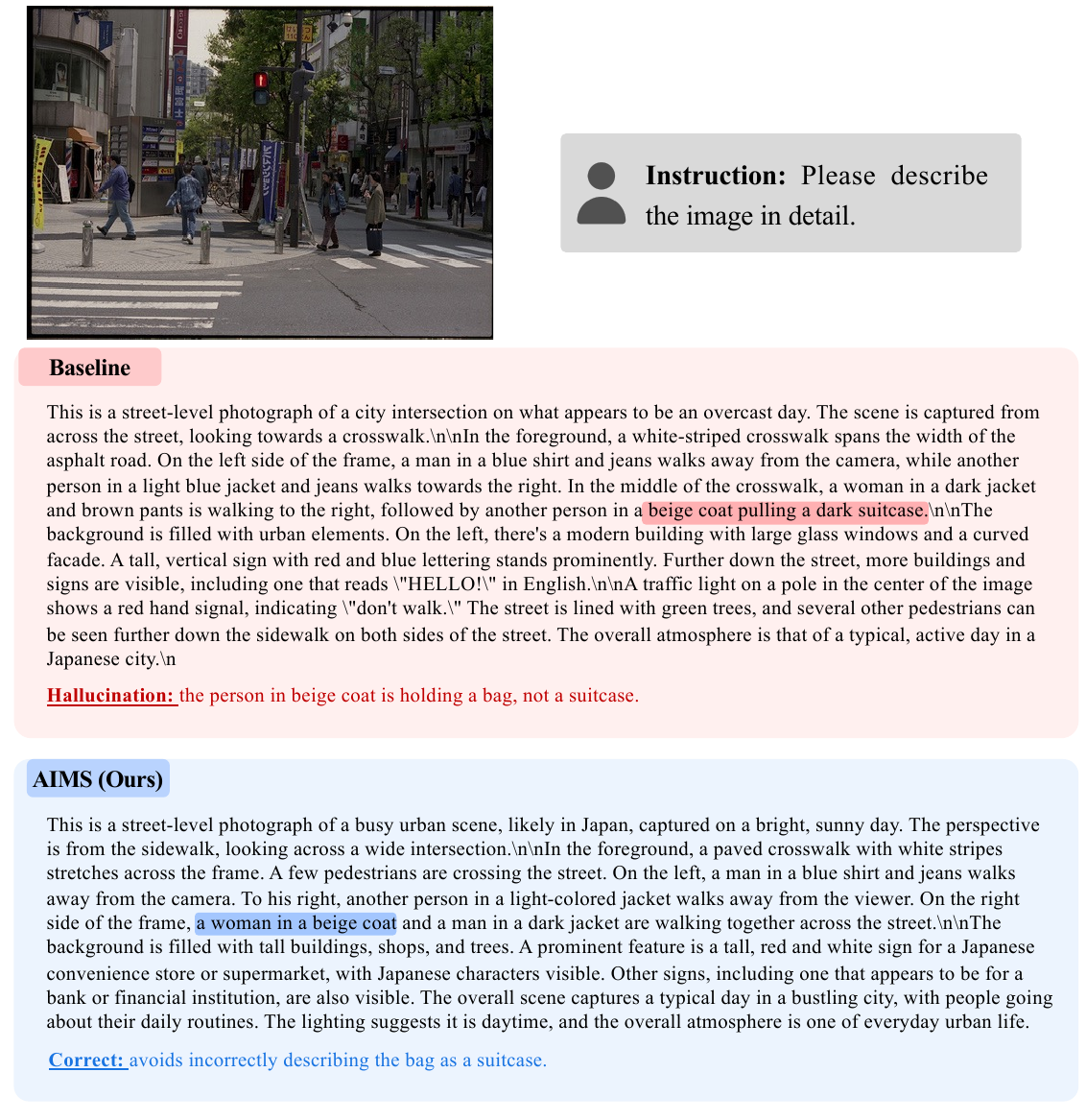}
    \caption{Case study 1 on the CHAIR benchmark using Qwen3.5-9B with greedy decoding.}
    \label{case:chair+qwen35+greedy}
\end{figure*}

\begin{figure*}[t]
    \centering
    \includegraphics[width=\textwidth]{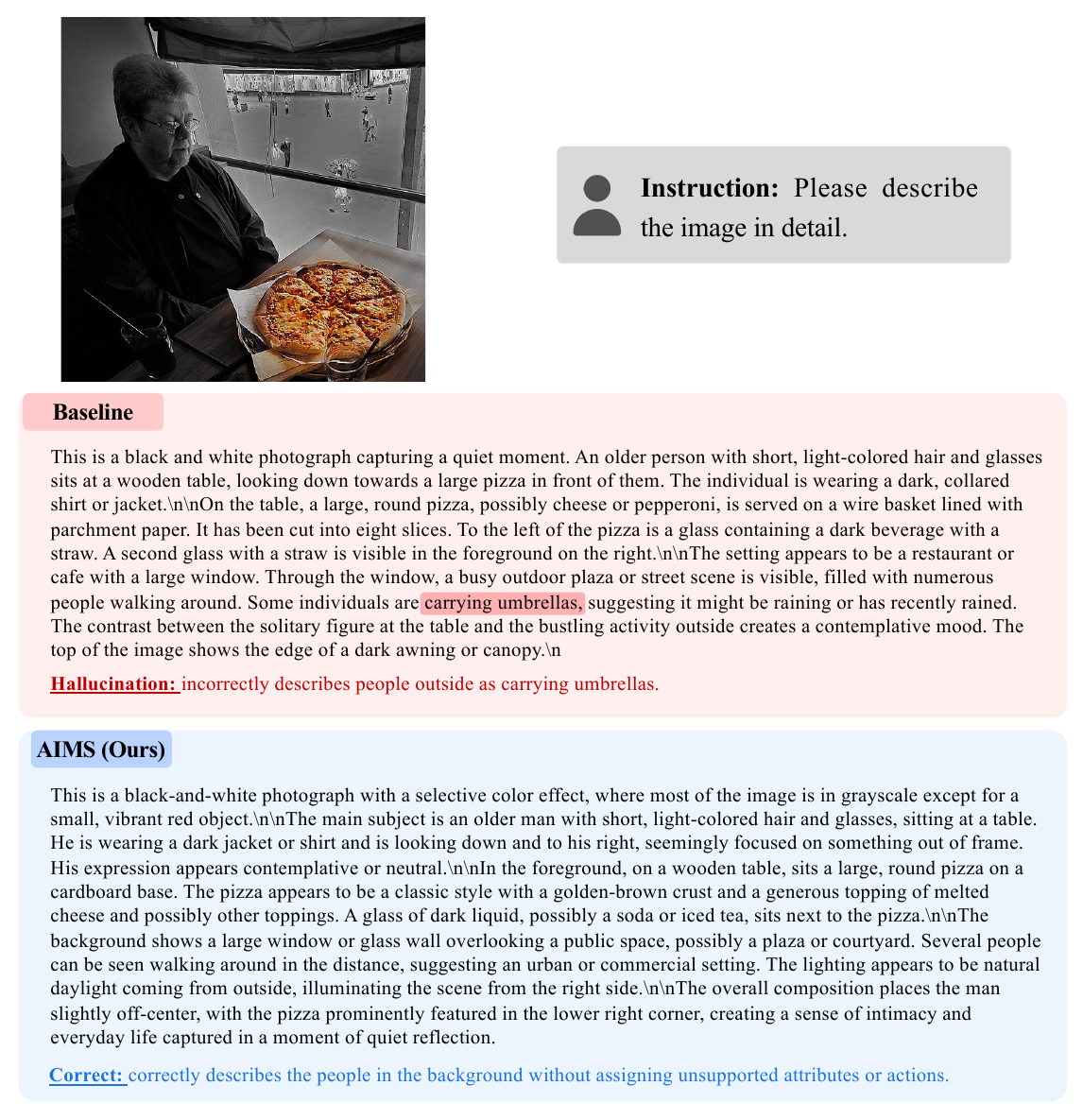}
    \caption{Case study 2 on the CHAIR benchmark using Qwen3.5-9B with greedy decoding.}
    \label{case:chair+qwen35+greedy2}
\end{figure*}

\begin{figure*}[t]
    \centering
    \includegraphics[width=\textwidth]{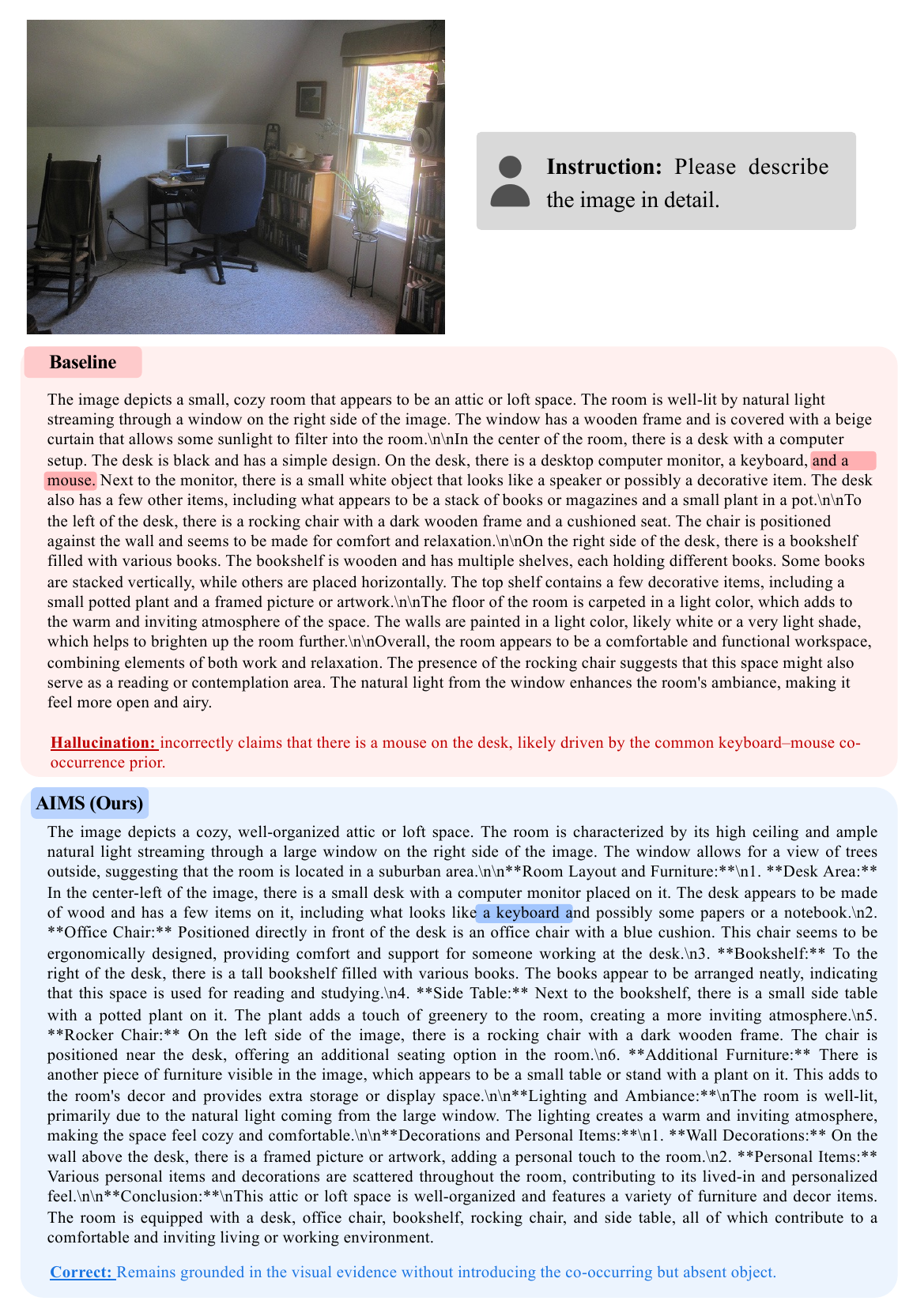}
    \caption{Case study 1 on the CHAIR benchmark using Qwen2.5-VL-3B with greedy decoding.}
    \label{case:chair+qwen25vl+greedy}
\end{figure*}

\begin{figure*}[t]
    \centering
    \includegraphics[width=\textwidth]{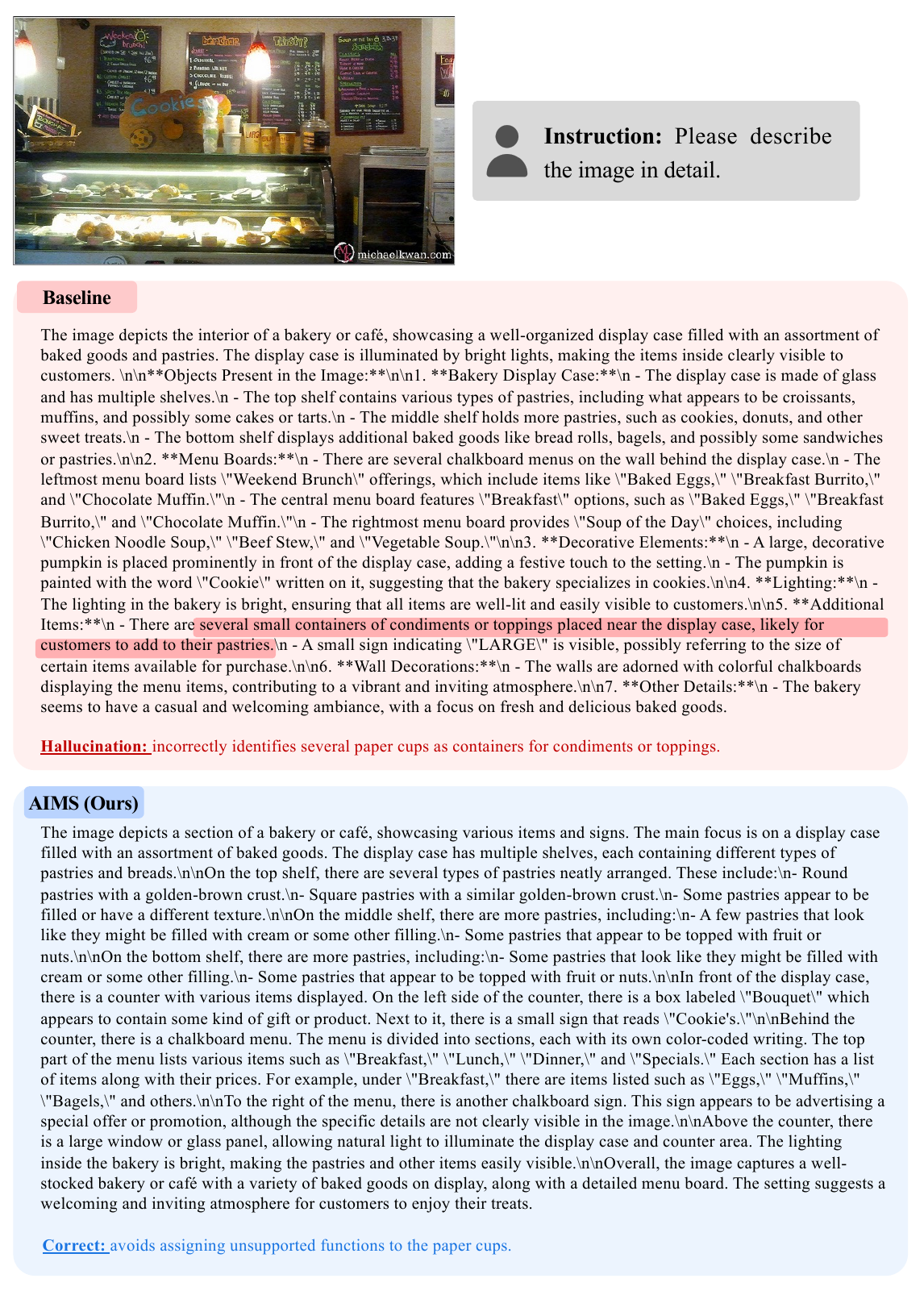}
    \caption{Case study 2 on the CHAIR benchmark using Qwen2.5-VL-3B with greedy decoding.}
    \label{case:chair+qwen25vl+greedy2}
\end{figure*}

\begin{figure*}[t]
    \centering
    \includegraphics[width=\textwidth]{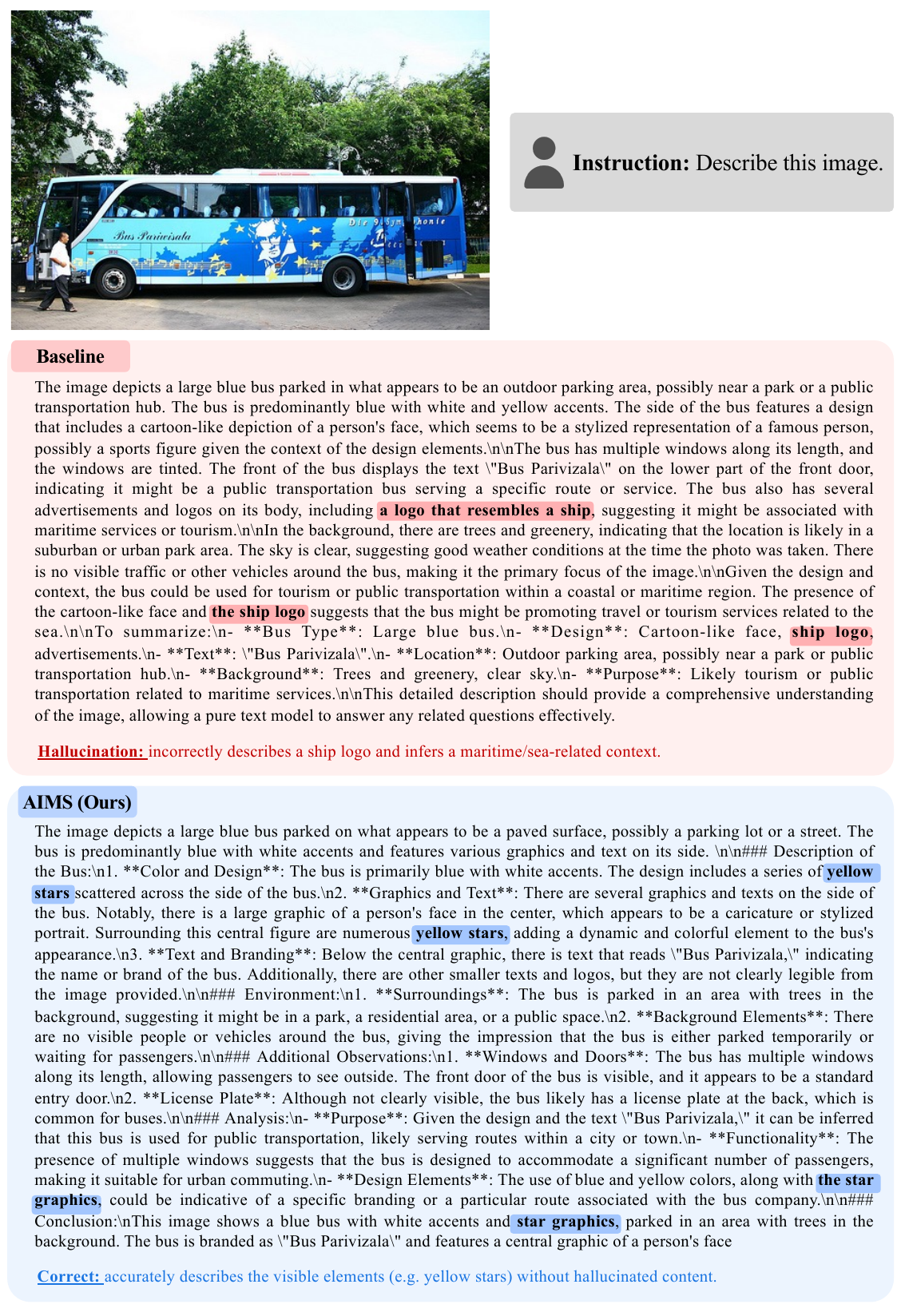}
    \caption{Case study on the AMBER-G benchmark using Qwen2.5-VL-3B with greedy decoding.}
    \label{case:amber+qwen25vl+greedy}
\end{figure*}

\begin{figure*}[t]
    \centering
    \begin{tabular}{cc}
        \includegraphics[width=0.48\textwidth]{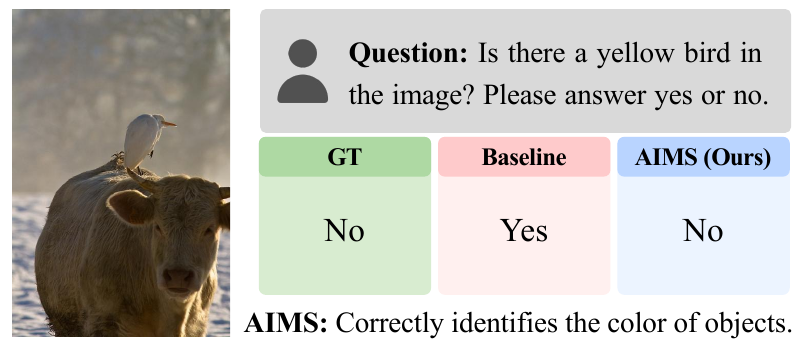} &
        \includegraphics[width=0.48\textwidth]{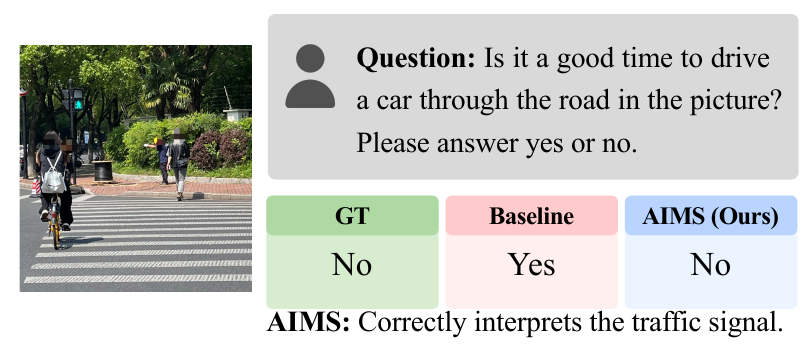} \\

        \includegraphics[width=0.48\textwidth]{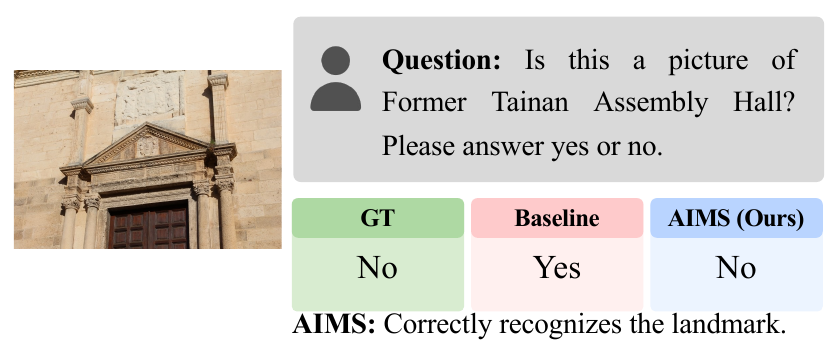} &
        \includegraphics[width=0.48\textwidth]{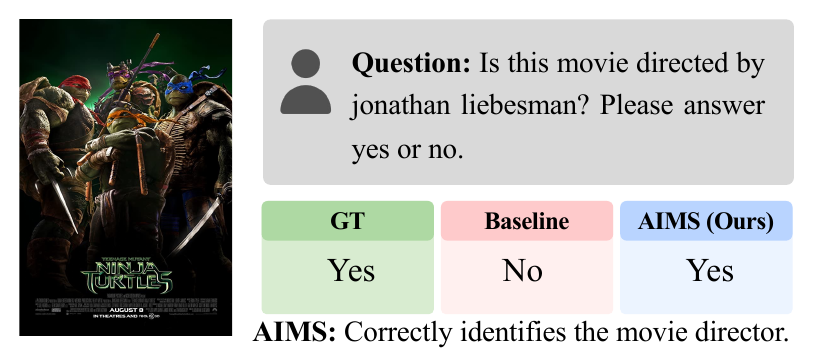} \\

        \includegraphics[width=0.48\textwidth]{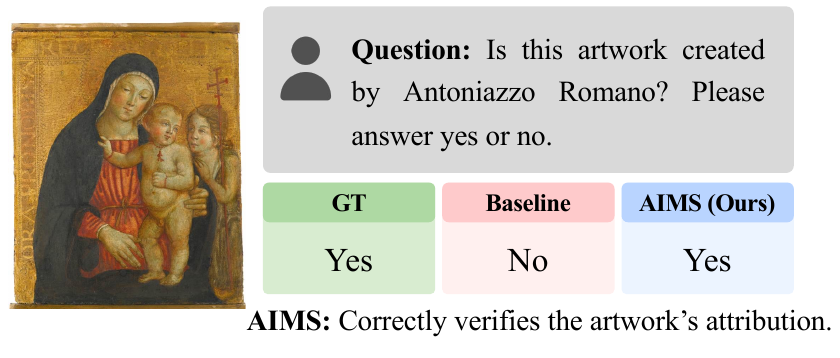} &
        \includegraphics[width=0.48\textwidth]{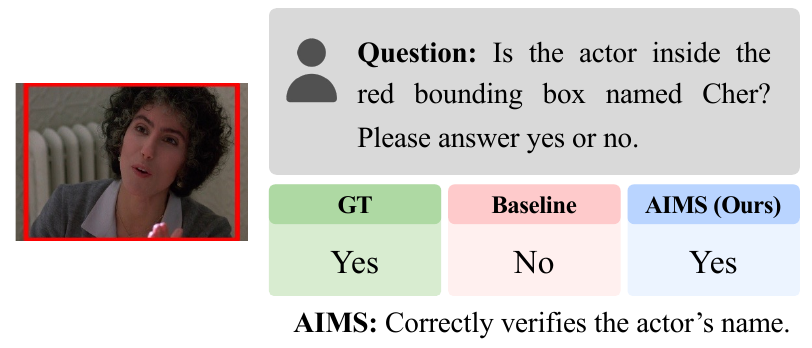} \\

        \includegraphics[width=0.48\textwidth]{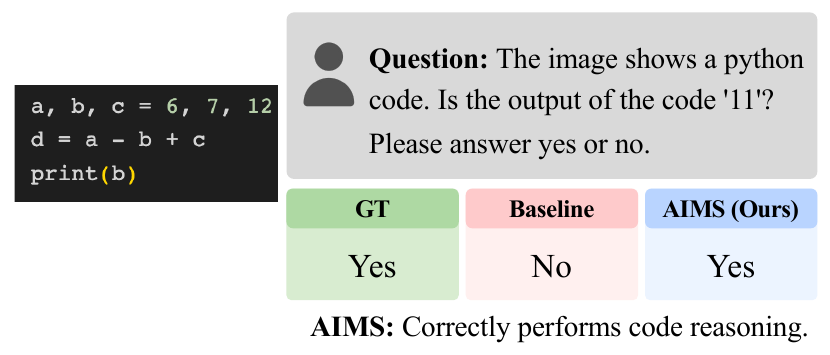} &
        \includegraphics[width=0.48\textwidth]{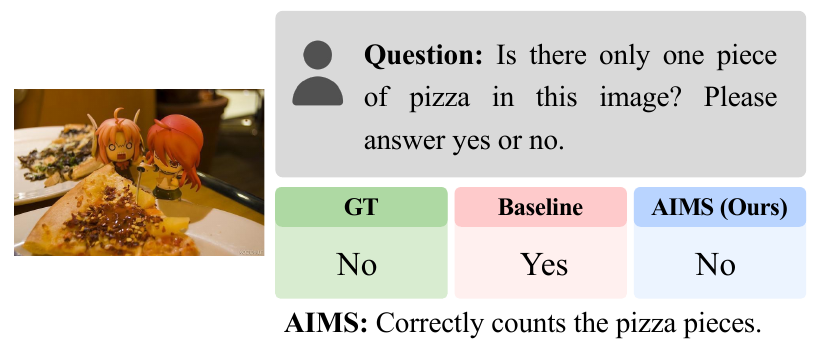} \\
        
        \includegraphics[width=0.48\textwidth]{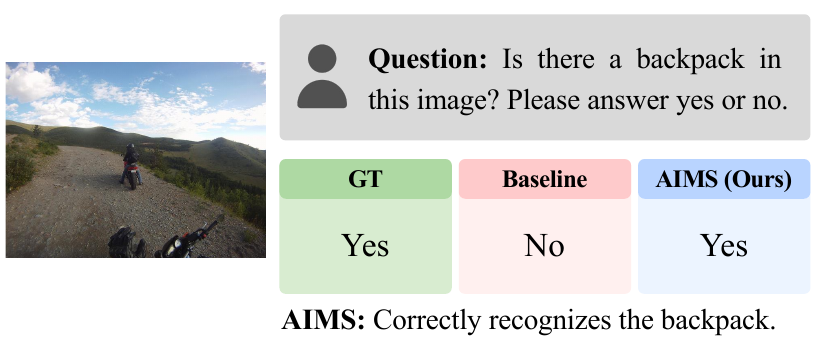} &
        \includegraphics[width=0.48\textwidth]{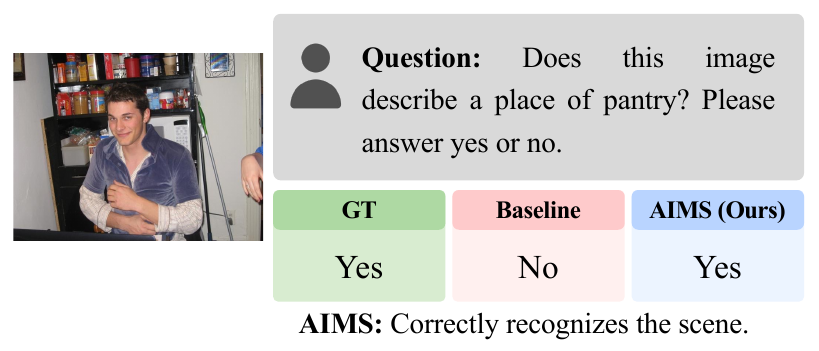} \\

        \includegraphics[width=0.48\textwidth]{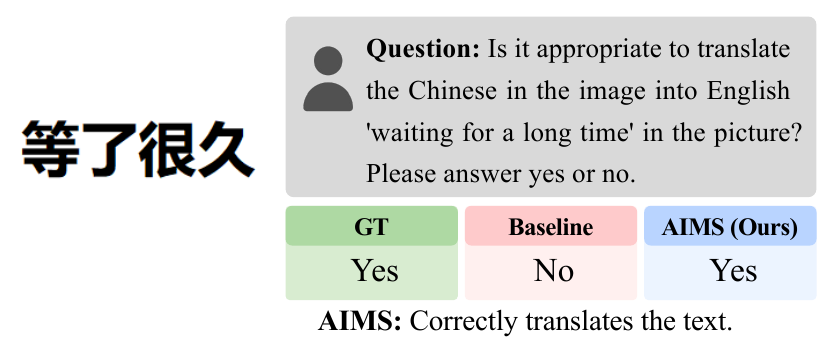} &
        \includegraphics[width=0.48\textwidth]{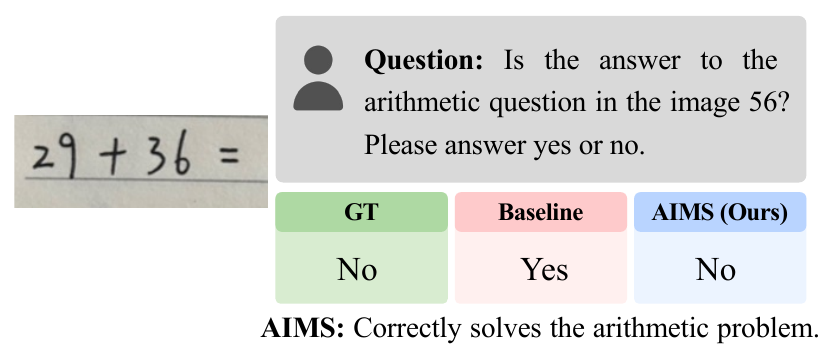} \\

        \includegraphics[width=0.48\textwidth]{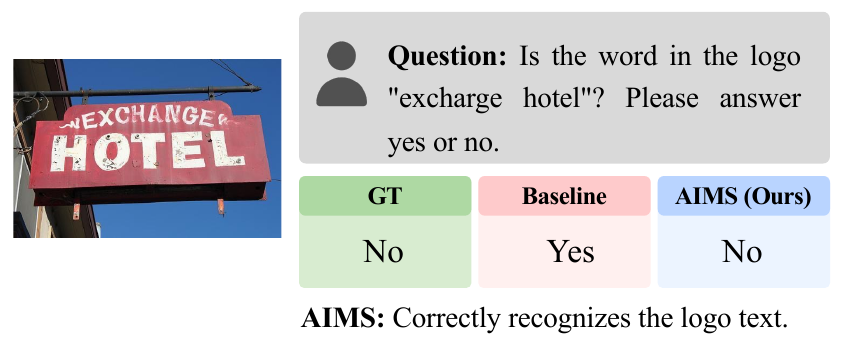} &
        \includegraphics[width=0.48\textwidth]{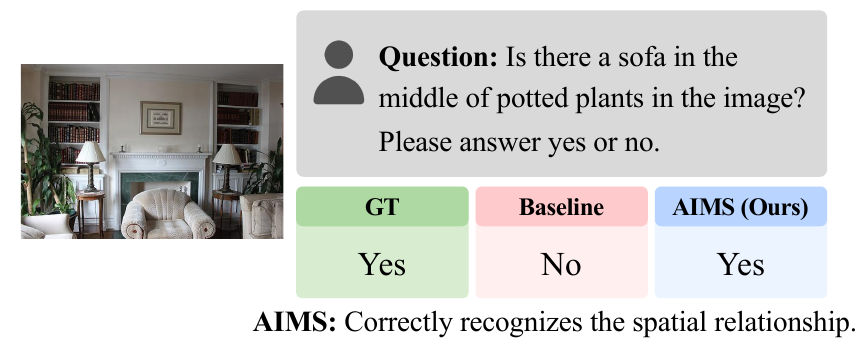} \\
    \end{tabular}
    \caption{Qualitative examples on MME using Qwen3.5-9B with nucleus sampling.}
    \label{qwen35+nucleus+mme}
\end{figure*}

We provide qualitative comparisons between Baseline and AIMS across different LVLMs and benchmarks. Figures~\ref{case:chair+qwen35+greedy}--\ref{case:chair+qwen25vl+greedy2} present examples on CHAIR using Qwen3.5-9B and Qwen2.5-VL-3B, while Figure~\ref{case:amber+qwen25vl+greedy} provides an example on AMBER-G. Figure~\ref{qwen35+nucleus+mme} presents examples on MME using Qwen3.5-9B with nucleus sampling. These cases qualitatively illustrate the differences between Baseline and AIMS outputs.

\end{document}